%% file: cvpr24.tex
\documentclass[10pt,twocolumn,letterpaper]{article}
\usepackage[accsupp]{axessibility} 

\usepackage{cvpr}              

\input{preamble}

\definecolor{cvprblue}{rgb}{0.21,0.49,0.74}
\usepackage[pagebackref,breaklinks,colorlinks,citecolor=cvprblue]{hyperref}

\usepackage{multirow}
\usepackage[utf8]{inputenc} 
\usepackage[T1]{fontenc}    
\usepackage{url}            
\usepackage{nicefrac}       
\usepackage{float}
\usepackage{microtype}
\usepackage{graphicx}
\usepackage{subcaption}
\usepackage{xcolor}
\usepackage{booktabs} 
\usepackage{lipsum}
\usepackage{xspace}
\usepackage{enumitem}
\usepackage{natbib}
\setcitestyle{numbers}
\setcitestyle{square}
\usepackage{booktabs} 
\usepackage{wrapfig}
\usepackage{makecell}

\usepackage{algorithm}
\usepackage{algpseudocode}

\usepackage{colortbl}
\usepackage{enumitem}
\usepackage{doi}

\usepackage{amsmath}
\usepackage{amssymb}
\usepackage{mathtools}
\usepackage{amsthm}

\usepackage[capitalize,noabbrev]{cleveref}

\usepackage[most]{tcolorbox}

\usepackage[most]{tcolorbox}
\usepackage{wrapfig}
\usepackage{xurl}

\input{macros}

\def\paperID{12817} 
\def\confName{CVPR}
\def\confYear{2024}

\title{
MetaCloak: Preventing Unauthorized Subject-driven Text-to-image Diffusion-based Synthesis via Meta-learning
}

\author{Yixin Liu$^{1,3}$\thanks{This work was done during Yixin's internship at Samsung Research America.} \quad Chenrui Fan$^{2}$ \quad 
Yutong Dai$^{1}$ \quad Xun Chen$^{3}$ \quad
Pan Zhou$^{2}$ \quad Lichao Sun$^{1}$ \\ 
\textsuperscript{1} Lehigh University 
\textsuperscript{2} Huazhong University of Science and Technology\\
\textsuperscript{3} Samsung Research America\\
{\tt\small yila22@lehigh.edu, \{fancr0422, panzhou\}@hust.edu.cn}\\
{\tt \small \{yud319, lis221\}@lehigh.edu, Xun.chen@samsung.com}
}

\begin{document}
\maketitle

\input{sections/abstract}
\input{sections/Introduction}

\input{sections/related_work}
\input{sections/preliminaries}
\input{sections/problem_statement}
\input{sections/method}

\input{sections/Experiments}

\input{sections/conclusion}
\newpage

{
    \small
    \bibliographystyle{ieeenat_fullname}
    \bibliography{ref}
}


\newpage
\appendix
\input{sections/appendix/details}
\input{sections/appendix/more-res}
\input{sections/appendix/recolor}

\input{sections/appendix/understanding}
\end{document}

%% file: preamble.tex
\usepackage[dvipsnames]{xcolor}


%% file: macros.tex
\newcommand{\secondrunner}[1]{\cellcolor{gray!25}#1}

\usepackage{amsmath,amsfonts,bm}

\def \header#1{\noindent {\bf #1.}}

\def\eqref#1{equation~\ref{#1}}

\def\1{\bm{1}}

\DeclareMathAlphabet{\mathsfit}{\encodingdefault}{\sfdefault}{m}{sl}
\SetMathAlphabet{\mathsfit}{bold}{\encodingdefault}{\sfdefault}{bx}{n}

\newcommand{\E}{\mathbb{E}}

\newcommand{\bc}{\mathbf{c}}

\newcommand{\bx}{\mathbf{x}}

\DeclareMathOperator*{\argmax}{arg\,max}
\DeclareMathOperator*{\argmin}{arg\,min}

\newcommand{\diffusion}{\hat{\mathbf{x}}}
\newcommand{\robLdb}{\mathcal{L}_\textrm{db}^\textrm{rob}}
\newcommand{\lcls}{\mathcal{L}_{\text{cls}}}

\newcommand{\promptOne}{{\scriptsize A photo of sks person}}
\newcommand{\promptTwo}{{\scriptsize A dslr portrait of sks person}}

%% file: sections/abstract.tex
\begin{abstract}
    Text-to-image diffusion models allow seamless generation of personalized images from scant reference photos. Yet, these tools, in the wrong hands, can fabricate misleading or harmful content, endangering individuals. To address this problem, existing poisoning-based approaches perturb user images in an imperceptible way to render them "unlearnable" from malicious uses. We identify two limitations of these defending approaches: i) sub-optimal due to the hand-crafted heuristics for solving the intractable bilevel optimization and ii) lack of robustness against simple data transformations like Gaussian filtering. To solve these challenges, we propose MetaCloak, which solves the bi-level poisoning problem with a meta-learning framework with an additional transformation sampling process to craft transferable and robust perturbation. Specifically, we employ a pool of surrogate diffusion models to craft transferable and model-agnostic perturbation. Furthermore, by incorporating an additional transformation process, we design a simple denoising-error maximization loss that is sufficient for causing transformation-robust semantic distortion and degradation in a personalized generation. Extensive experiments on the VGGFace2 and CelebA-HQ datasets show that MetaCloak outperforms existing approaches. Notably, MetaCloak can successfully fool online training services like Replicate, in a black-box manner, demonstrating the effectiveness of MetaCloak in real-world scenarios. Our code is available at \href{https://github.com/liuyixin-louis/MetaCloak}{https://github.com/liuyixin-louis/MetaCloak}. 
\end{abstract}

%% file: sections/Introduction.tex
\section{Introduction}
\label{sec:intro}
Diffusion models achieve significant success in a wide range of applications, including image generation \citep{ho2020denoising, song2021scorebased, dhariwal2021diffusion}, image editing \citep{kim2022diffusionclip, shi2023dragdiffusion, choi2023customedit}, and text-to-image synthesis \citep{rombach2022highresolution}. Subject-driven text-to-image synthesis, an emerging application of diffusion models, in particular, has attracted considerable attention due to its potential to generate personalized images from a few reference photos. 
Among the approaches proposed to achieve this goal \citep{ramesh2022hierarchical, saharia2022photorealistic}, DreamBooth \citep{ruiz2023dreambooth} and Text Inversion \cite{gal2022image} are two prominent training-based methods that offer impressive personalize generation ability. With an additional lightweight model fine-tuning process for capturing the subject or a concept-related embedding training phase, personalized diffusion models can retain the generation capacity from the pre-training stage and conduct vivid personalized generation. While these methods empower high-quality personalized generation, they also raise privacy concerns as they can fabricate misleading or harmful content in the wrong hands, endangering individuals. For example, recent news \citep{npr,thenextweb,totonews} indicates that AI tools like diffusion models have been employed to generate fake profiles of individuals for launching a new wave of fraud.

\begin{figure}
    \centering
    \includegraphics[width=0.9\linewidth]{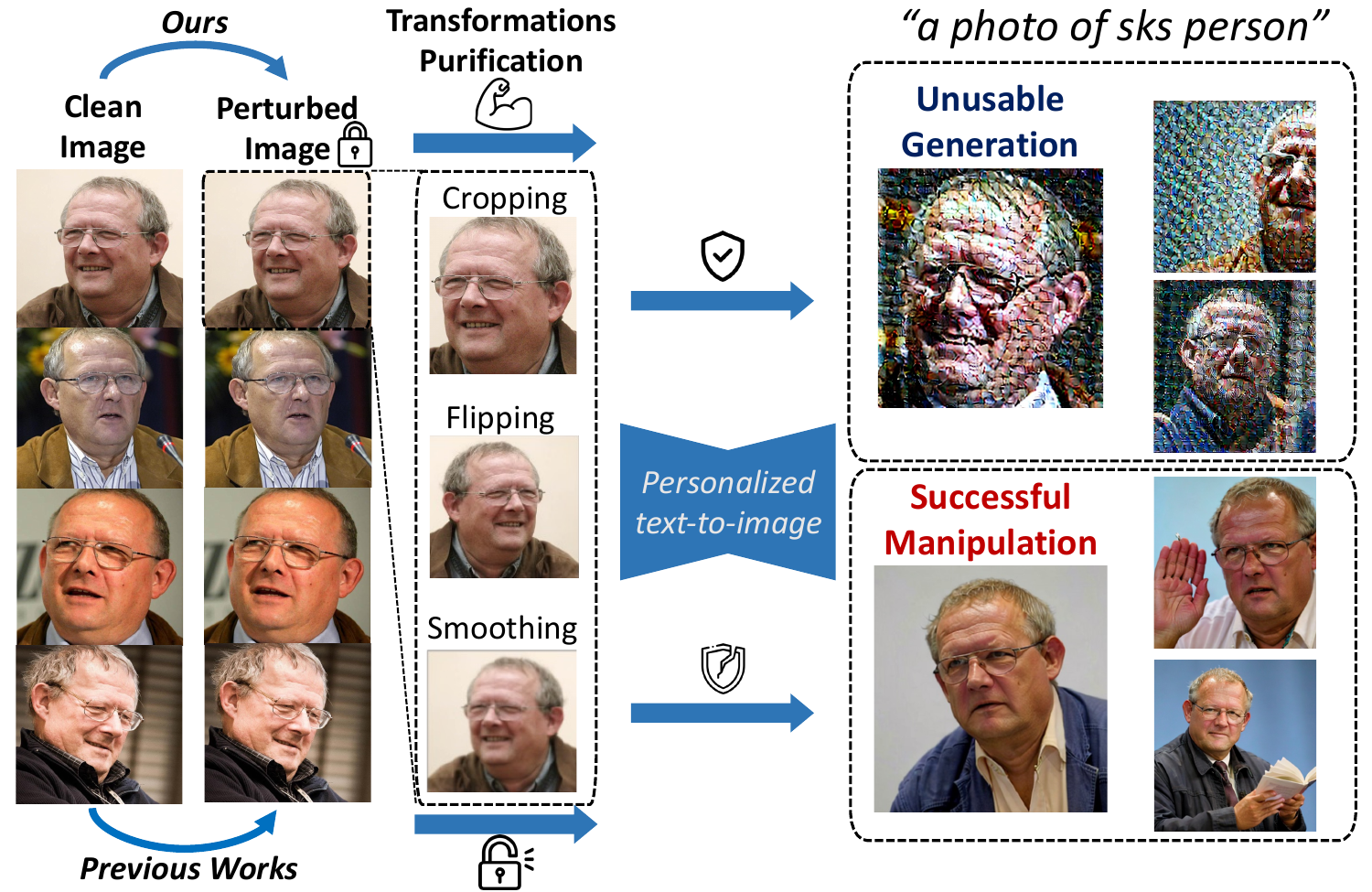}
    \caption{Image protected by existing methods fails to fool personalized text-to-image approaches after applying data transformations. In contrast, our MetaCloak is still robust in such adversity. 
    }
    \label{fig:motivation}
\end{figure}

To tackle these issues, some poisoning-based approaches \citep{le_etal2023antidreambooth, liang2023adversarial} have been recently proposed to perturb user images in an imperceptible way to render them ``unlearnable'' from malicious uses. Specifically, these approaches aim to craft perturbations that can mislead the personalized diffusion model generation process to compromise the generation ability. 
For instance, PhotoGuard \cite{salman2023raising} proposes to attack the VAE encoder or UNet decoder for crafting perturbation that aligns the perturbed latent code or denoised image to the ones of a dummy target. Similarly, \citet{liang2023adversarial} proposes AdvDM to craft protected images with a pre-trained fixed surrogate model with adversarial perturbation. Its latter version, Mist \citep{liang2023mist}, considers an additional targeting loss for degrading the texture quality. Targeting art style transferring, Glaze \cite{shan2023glaze} proposes to minimize the feature distance between the perturbed image and a targeting image with a pre-trained feature extractor. However, these works all focus on attacking text-to-image synthesis methods that leverage \textit{fixed} diffusion models, where protection is much easier due to the inevitable adversarial vulnerability of the DNNs. 

In contrast, fine-tuning-based personalized generation approaches, represented by DreamBooth, ensure high-quality synthesis via \textit{actively learning} the concept of the subject by optimizing either the model parameters or a concept-related embedding. 
Compared to attacking \textit{fixed diffusion model}, success in attacking these fine-tuning-based generation methods is rooted in different mechanisms, i.e., tricking the models into establishing false correlations or overfitting. 
To the best of our knowledge, Anti-Dreambooth \citep{le_etal2023antidreambooth} is the only existing work focused on this challenging setting. Although this approach has shown effectiveness in preventing unauthorized subject-driven text-to-image synthesis, it exhibits two limitations. 
Firstly, it uses hand-crafted heuristics to address the underlying poisoning problem, which is a challenging bilevel optimization problem, yielding sub-optimal performance. Specifically, it incorporates an additional clean set training process in the surrogate model training trajectory, which is mismatched with the actual ones trained on the poison and thus degrades the poisoning capacity. 
Secondly, recent studies found that these data protections are fragile and demonstrate limited robustness against minor data transformations such as filtering (see Fig. \ref{fig:motivation}). Given these limitations, in this work, we ask the following question: \textit{Can we design a more effective and robust data protection scheme that can prevent unauthorized subject-driven text-to-image diffusion-based synthesis under data transformation?}

To answer this question, we propose MetaCloak, a more effective and robust poisoning-based data protection approach against unauthorized subject-driven text-to-image diffusion-based synthesis. To tackle the first challenge, as well as to improve perturbation transferability, MetaCloak leverages meta-learning to learn model-agnostic perturbation over a pool of steps-staggered surrogates. 
With this meta-learning framework, we found that the perturbations are more effective than the previous approaches under the training setup without data transformations. 
To address the second challenge, i.e., improving the robustness of perturbation under the training setting with data transformation, we incorporate a transformation sampling process to craft transformation-robust perturbation. 
Besides, designing generation loss to guide poisoning is tricky since leveraging the ``ground-truth'' metrics might lead to an over-fitting problem. To address this problem, we design a denoising-error maximization loss to encourage the added perturbation to cause serious semantic distortion from the model perception, which aims to introduce ``meaningless'' and ``hard-to-understand'' patterns that can trick the model into learning false correlations. 
On the VGGFace2 and CelebA-HQ datasets, we show that our method significantly outperforms existing approaches under data transformations. Notably, MetaCloak can successfully fool online personalized diffusion model training services like Replicate in a black-box manner, demonstrating the effectiveness of MetaCloak in real-world scenarios. Our main contributions are summarized as follows. \looseness=-1

\begin{enumerate}
    \item We propose MetaCloak, the first robust poisoning-based approach for crafting more effective perturbation that can further bypass data transformation defenses.
    \item To solve the intractable bi-level poisoning, we propose to use meta-learning to learn model-agnostic and transferable perturbation by leveraging a pool of surrogates. 
    \item To guide the poisoning and bypass transformation defenses, we design a simple denoising-error maximization loss with an additional transformation process for transformation-robust semantic distortion. 
    \item Extensive experiments on the VGGFace2 and CelebA-HQ datasets show the superiority of our method compared to existing approaches under settings with and without data transformations. 
    Notably, our method can fool online training services like Replicate in a black-box manner. 
\end{enumerate}

%% file: sections/related_work.tex
\section{Related Wroks}
\label{sec. related work}

\header{Availability Attack}
Availability attack aims to degrade data availability for model training by injecting imperceptible perturbations into the training data. The underlying mechanism of this attack is rooted in the shortcut learning preference of deep neural networks (DNNs), where models turn to learn easy but brittle features for prediction. 
Existing works on avability attacks mainly focus on image classification \cite{huang2021unlearnable, huang2020metapoison, feng2019learning, yu2022availability}. Unlearnable Example \cite{huang2021unlearnable} and L2C \cite{feng2019learning} are two of the pioneers in this field, and later research \citet{tao2021better} indicates that noise can be easily removed with adversarial training and data transformations. To address this concern, many works have been proposed, with techniques of min-max optimization \cite{fu2021robust}, filter-based perturbation \cite{sadasivan2023cuda}, and score guiding \cite{fang2023towards}. To better understand the defense mechanism, \citet{tao2021better} propose frameworks for showing some certificated purification against such attacks. To adapt to other models and applications, studies are also conducted in natural language processing \cite{ji2022unlearnable}, graph learning \cite{liu2023graphcloak}, and contrastive learning \cite{ren2022transferable}. 
To the best of our knowledge, we are the first to study robust availability attacks to personalized generation with diffusion models. 

\header{Protection Against Unauthorized Subject-driven AI Synthesis}
Unauthorized subject-driven AI synthesis including style transfer \cite{kim2022diffusionclip}, personalized generation \cite{ruiz2023dreambooth}, and image inpainting \cite{bar2022visual}, 
pose a serious threat to the privacy of individuals and art copyright \cite{carlini2023extracting, vyas2023provable}. To address this concern, recent works \cite{salman2023raising,zheng2023understanding, shan2023glaze, le_etal2023antidreambooth} propose to protect the data with poisoning-based approaches, which modify the data in an imperceptible way while causing severe degradation in the generation performance. PhotoGuard \cite{salman2023raising} proposes to attack the VAE encoder or UNet decoder for crafting perturbation that aligns the perturbed latent code or denoised image to the ones of the dummy target. Targeting degrading art style transferring, Glaze \cite{shan2023glaze} proposes to minimize similar targeting loss with a pre-trained style-transfer model, with learned perceptual similarity as a penalty. Similarly, AdvDM \cite{zheng2023understanding} 
proposes to craft perturbation that minimizes the likelihood of the perturbed image with a pre-trained diffusion model, which equivalently maximizes the denoising loss during training. 
Its later version, Mist \cite{liang2023mist}, further incorporates a texture targeting loss for a more robust and sharp pattern into the perturbed images. However, these works all focus on attacking the synthesis methods that leverage \textit{fixed} diffusion models by exploiting the adversarial vulnerability of the DNNs. In contrast, our work focuses on conducting more challenging data protection against fine-tuning-based synthesis methods \cite{ruiz2023dreambooth, raj2023dreambooth3d}. To the best of our knowledge, Anti-DreamBooth \cite{le_etal2023antidreambooth} is the only work that studies data protection against fine-tuning-based synthesis approaches.
However, its protection is brittle under transformation purification. In this work, we study poisoning-based data protection with better effectiveness and robustness against transformations. 

%% file: sections/preliminaries.tex
\section{Preliminary}
\label{sec:preliminary}
\label{sec:t2i}
\noindent \textbf{Text-to-Image Diffusion Models.}
Diffusion models are probabilistic generative models that are trained to learn a data distribution by the gradual denoising of a variable sampled from a Gaussian distribution. 
Our specific interest lies in a pre-trained text-to-image diffusion model denoted as $\diffusion_\theta$. This model operates by taking an initial noise map $\boldsymbol{\epsilon}$ sampled from a standard Gaussian distribution $\mathcal{N}(\mathbf{0}, \mathbf{I})$ and a conditioning vector $\mathbf{c}$. This conditioning vector $\mathbf{c}$ is generated through a series of steps involving a text encoder represented as $\Gamma$, a text tokenizer denoted as $f$, and a text prompt $\mathbf{P}$ (i.e., $\mathbf{c}=\Gamma(f(\mathbf{P}))$). The ultimate output of this model is an image denoted as $\mathbf{x}_{\text{gen}}$, which is produced as a result of the operation $\mathbf{x}_{\text{gen}}=\diffusion_\theta(\boldsymbol{\epsilon}, \mathbf{c})$.
They are trained using a squared error loss to denoise a variably-noised image
as follows:
\begin{equation}
\label{eq:denoise}
    \mathcal{L}_{\text{denoise}}(\bx, \bc; \theta)=\mathbb{E}_{\boldsymbol{\epsilon},t}[{w_t \|\diffusion_\theta(\alpha_t \mathbf{x} + \sigma_t \boldsymbol{\epsilon}, \mathbf{c}) - \mathbf{x} \|^2_2}], 
\end{equation}
where $\mathbf{x}$ is the ground-truth image, $\mathbf{c}$ is a conditioning vector (e.g., obtained from a text prompt), and $\alpha_t, \sigma_t, w_t$ are terms that control the noise schedule and sample quality, and are functions of the diffusion process time $t$. 

\noindent \textbf{Adversarial Attacks to Text-to-Image Diffusion Models.}
\label{sec:advt2i}
Adversarial attacks aim to perform an imperceptible perturbation on the input image in order to mislead machine learning models' predictions. 
In the classification scenario, for a given classifier $f_{\text{cls}}$, a perturbed adversarial image $\bx'$ is generated from the original image $\bx$ to misguide the model into incorrect classification. Constraints on the perceptibility of changes are often imposed through $\ell_p$ norms (with $p\geq 1$), such that the perturbed image $\bx^\prime$ is bounded within a $\ell_p$-ball centered at $\bx$ with radius $r>0$, i.e., $x^\prime \in B_p(\bx, r)=\left\{\bx^\prime: \Vert \bx' - \bx\Vert_p \leq r \right\}$. Given a classification loss $\lcls$, untargeted adversarial examples are crafted by solving $\max_{\bx^\prime \in B_p(\bx, r)} \lcls(f_{\text{cls}}(\bx^\prime), y_\text{true})$, where $y_\text{true}$ is the true label of image $\bx$. For the text-to-image generation scenario, given a pre-trained text-to-image diffusion model $\diffusion_\theta$, the adversarial attack aims to perturb the image to hinder the model from reconstructing the image, i.e., $\bx^\prime \gets \argmax_{\bx^\prime \in B_p(\bx, r)} \mathcal{L}_{\text{denoise}}(\bx^\prime, \bc; \theta)$.
In this paper, we consider the $\ell_\infty$-norm for its alignment with perception \cite{goodfellow2014explaining}. To solve this constrained optimization, the Projected Gradient Descent (PGD) \citep{madry2018towards} technique is commonly utilized by iteratively updating the poisoned image $\bx'$. Formally, the adversarial example $\bx'$ is updated as
\begin{align}
    \bx^\prime_{i} = \Pi_{B_\infty(\bx,r)}(\bx^\prime_{i-1} + \alpha \mathrm{sign}(\nabla_{\bx_{i-1}'}  \mathcal{L}_{\text{denoise}})),
\label{eq:pgd_upd}
\end{align}
where $\bx_0'=\bx$, $\text{sign}(\cdot)$ is the sign function, $i$ is the step index, and the step size $\alpha>0$. During this generation process, the adversarial examples gradually progress in a direction that would increase the denoising loss while maintaining imperceptible perturbations. Recent works \cite{le_etal2023antidreambooth, liang2023mist} have demonstrated that perturbed images from this attack can effectively deceive text-to-image generation models \citep{gal2022image, ruiz2023dreambooth} to produce low-quality images.

\noindent \textbf{Personalized Diffusion via DreamBooth Fine-turning.}
DreamBooth is a method aimed at personalizing text-to-image diffusion models for specific instances. It has two main objectives: first, to train the model to generate images of the given subject with generic prompts like ``a photo of \textit{sks} [class noun]", where \textit{sks} specifies the subject and ``[class noun]" is the category of object (e.g., ``person"). For this, it uses the loss defined in Eq.~\ref{eq:denoise} with $x_u$ as the user's reference image and conditioning vector $\bc:= \Gamma(f(\text{``a photo of \textit{sks} [class noun]"}))$. Similar to the classification model, this 
guides the model to create the correlation between the identifier and the subject. Secondly, it introduces a class-specific prior-preserving loss to mitigate overfitting and language-drifting issues. Specifically, it retains the prior by supervising the model with \textit{its own generated samples} during the fine-tuning stage. With a class-specific conditioning vector $\bc_\text{pr} := \Gamma(f(\text{`` photo of a [class noun]"}))$ and random initial noise $\mathbf{z}_{t_1} \sim \mathcal{N}(\mathbf{0}, \mathbf{I})$, DreamBooth first generates prior data $\mathbf{x}_{\mathrm{pr}}=\diffusion_{\theta_0}\left(\mathbf{z}_{t_1}, \mathbf{c}_{\mathrm{pr}}\right)$ using the pre-trained diffusion model and then minimize: 
\begin{equation}
    \label{eq:DB_loss}
    \begin{aligned}
        \mathcal{L}_{\text{db}}(\bx, \bc; \theta)=
  \mathbb{E}_{\boldsymbol{\epsilon}, \boldsymbol{\epsilon}^{\prime}, t}\left[w_t\left\|\diffusion_\theta\left(\alpha_t \mathbf{x}+\sigma_t \boldsymbol{\epsilon}, \mathbf{c}\right)-\mathbf{x}\right\|_2^2+\right. \\
    \left.\lambda w_{t^{\prime}}\left\|\diffusion_\theta\left(\alpha_{t^{\prime}} \mathbf{x}_{\mathrm{pr}}+\sigma_{t^{\prime}} \boldsymbol{\epsilon}^{\prime}, \mathbf{c}_{\mathrm{pr}}\right)-\mathbf{x}_{\mathrm{pr}}\right\|_2^2\right],
    \end{aligned}
    \end{equation}
where $\epsilon, \epsilon'$ are both sampled from $\mathcal{N}(0, \mathbf{I})$, the second term is the prior-preservation term that supervises the model with its own generated images and $\lambda$ controls for the relative importance of this term. 
With approximately one thousand training steps and four subject images, it can generate vivid personalized subject images with Stable Diffusion \citep{von-platen-etal-2022-diffusers}.

%% file: sections/problem_statement.tex
\section{Problem Statement}
\label{sec:problem}
We formulate the problem as follows. \textit{A user (image protector)} wants to protect his images ${X}_c = \{\bx_i\}_{i=1}^{n}$ from being used by an \textit{unauthorized model trainer} for generating personalized images using DreamBooth, where $n$ is the number of images. To achieve this, for some portion of images $\bx \in X_c$, the user injects a small perturbation onto the original image to craft poisoned images set $ X_p = \{\bx_i^\prime\}_{i=1}^{n}$, which is then published to the public. Later, the model trainers will collect and use $X_p$ to finetune a text-to-image generator $\diffusion_\theta$, following the DreamBooth algorithm, to get the optimal parameters $\theta^*$. We assume that the model trainer is aware of the poisoning to some extent, so some data transformations like filtering or cropping might be applied to the training image set $X_p$ during the data pre-processing phase of DreamBooth training. 
The objective of the user is to craft a delusive and robust image set $X_p$ to degrade the DreamBooth's personalized generation ability, which can be formulated as:
\begin{equation}
    X_p^* \in \argmax_{X_p, \theta^*}  \mathcal{L}_{\text{gen}}^* (X_{\text{ref}};\diffusion_{\theta^*}, X_p) \label{eq:bilevel.upper}
\end{equation}
\begin{equation}
\begin{split}
    \text{s.t. }\quad  \theta^* \in \argmin_{\theta} \{&\mathcal{L}_{\text{db}}^{\text{rob}}(X_p, T; \theta):= \\
    &\mathbb{E}_{\bx_i'\sim X_p, g\sim T } 
      \mathcal{L}_{\text{db}}(g(\bx_i' ), \bc;\theta)\}.
\end{split}
\label{eq:bilevel.lower}
\end{equation}

Here, $\bc$ is the class-wise conditional vector, $\mathcal{L}_{\text{gen}}^*$ is some perception-aligned loss to measure the personalization generation ability of trained model $\diffusion_{\theta^*}$ (with more details in the next section), $T$ is a set of data transformations the expected adversary might use, $X_{\text{ref}}$ is a clean reference set, and $\bc$ is the conditioning vector. Compared to vanilla $\mathcal{L}_{\text{db}}$ in (\ref{eq:DB_loss}), $\mathcal{L}_{\text{db}}^{\text{rob}}$ \footnote{We by default omit $T$ and simplify the notation as $\mathcal{L}_{\text{db}}^{\text{rob}}(X_p; \theta)$ in the following context. } is more robust to learning personalized diffusion models.

\header{Overall Goals} While it's hard to quantify a unified evaluation loss $\mathcal{L}_{\text{gen}}^*$ to measure the personalized generation quality, our overall goal is to degrade the usability of generated images, and we attempt to decompose the evaluation metric into the following two aspects: \textit{quality-related and semantic-related distortion}. Specifically, we seek to render the generated image awful quality by tricking the victim's model into generating an image with some artifacts. With this distortion, the model trainer can't use those images for some quality-sensitive applications. Furthermore, the subject identity of generated images should be greatly distorted for other's utilization. We'll dive into the design of $\mathcal{L}_{\text{gen}}^*$ in Sec. \ref{sec. loss}. 

%% file: sections/method.tex
\section{Method}
\label{sec:method}
\subsection{
Learning to Learn Transferable and Model-agnostic Perturbation
}
\label{sec:meta-learning}
\vspace{-4pt}

One naive idea to solve the bilevel problem~(\ref{eq:bilevel.upper})-(\ref{eq:bilevel.lower}) is to unroll all the training steps and optimize the protected examples $X_p$ via backpropagating. However, accurately minimizing this full bi-level objective is intractable since a computation graph that explicitly unrolls $10^3$ SGD steps would not fit on most of the current machines. To address this issue, inspired by~\citet{huang2020metapoison},
we propose to approximately optimize the upper-level objective~(\ref{eq:bilevel.upper}) and lower-level objective~(\ref{eq:bilevel.lower}) in an alternative fashion.
Specifically, considering the $i$-th iteration, when the current model weight $\theta_i$ and the protected image set $X_P^{i}$ are available (with $\theta_0$ is initialized from pre-trained diffusion model and $X_P^{0}=X_c$), we make a copy of current model weight $\theta_{i,0}^\prime\gets\theta_i$ for noise crafting and optimize the lower-level problem for $K$ steps as:
\begin{equation}
    \theta^\prime_{i,j+1} = \theta^\prime_{i, j} - \beta\nabla_{\theta_{i,j}'}\robLdb(X_p^{i}; \theta_{i,j}'), 
    \label{eq:unroll1}
\end{equation}
where $j \in \{0, 1, \dots, K-1\}$ and $\beta>0$ is the stepsize. We term this procedure $K$-step method. This unrolling procedure allows us to ``look ahead'' in training and view how the perturbations \textit{now} will impact the generation loss $\mathcal{L}_{\text{gen}}^*$ \textit{after} $K$ steps. We then leverage the unrolled model $\hat{\bx}_{\theta^\prime_{i,K}}$ for optimizing the upper-level problem, i.e., updating the protected images $X_p$ as:
\begin{equation}
    \scriptsize 
     X_{p}^{i+1} = \Pi_{B_\infty(X_{p}^{0},r)}(X_{p}^{i} + \alpha \mathrm{sign}(\nabla_{X_{p}^{i}} \mathcal{L}_{\text{gen}}^*(X_{\text{ref}}; \hat{\bx}_{\theta^\prime_{i,K}}, X_p^{i})). 
    \label{eq:unroll-example}
\end{equation}
After obtaining the updated protected images $X_p^{i+1}$, the surrogate model $\theta_{i}$ is trained with $\robLdb$ for a few SGD steps on $X_p^{i+1}$ to get $\theta_{i+1}$ as the next iteration's starting point
\begin{align}
    \theta_{i+1} = \theta_{i} - \beta\nabla_{\theta_{i}}\robLdb(X_p^{i+1}; \theta_{i}). 
    \label{eq:advanced}
\end{align}
The procedure (\ref{eq:unroll1})-(\ref{eq:advanced}) is executed repeatedly until the surrogate model reaches maximum training steps to obtain the final protected images $X_p^*$.
While this $K$-step method offers satisfactory results, it is not robust under various training settings with different models and initialization. Since training with a single surrogate model will turn to overfit on a single training trajectory \cite{huang2020metapoison}.
To craft transferable perturbations, we treat the injected perturbations as a training ``hyperparameter'', and use meta-learning to learn model-agnostic perturbations that degrade the performance of trained models. Different from the conventional meta-learning setting, whose goal is to transfer across datasets and tasks, in our poisoning problem, we aim to craft perturbations that transfer across different models and different training trajectories. To achieve this, we propose to learn perturbation over a pool of steps-staggered surrogates. 
Specifically, 
we train $M$ surrogate models $\{\theta^{j}\}_{j=1}^{M}$ as the initial point for the $j$-th surrogate model. Given maximum training steps $N_\textrm{max}$, the $j$-th surrogate model $\theta^{j}$ is trained with $\robLdb$ for $\lfloor jN_\textrm{max}/M \rfloor $ steps \textit{on clean data} from the pre-trained weight $\theta_0$. With this pool, we then conduct $C$ crafting outer loop; for each loop, we sample a batch of surrogates and train them separately on the poison. Upon the completion of inner-loop training, we average the generation losses $\mathcal{L}_{\text{gen}}^*$ over ensembles and update the protected images $X_p$ with gradient ascent. Following MAML \citep{finn2017model}, a first-order gradient approximation is used for efficient perturbation crafting.

\subsection{Transformation-robust Semantic Distortion with Denoising-error Maximization}
\label{sec. loss}
During the evaluation stage of the generated images, we can readily leverage various quality reference-based and reference-free assessment metrics like CLIP-IQA \citep{wang2022exploring} and FDSR \citep{he2021fast} for the construction of the ground-truth generation loss $\mathcal{L}_{\text{gen}}^*$. However, during the poisoning stage, we can not simply take these ``ground-truth'' metric losses to serve as the loss for crafting noise: i) overfitting is prone to happen since most quality-assessment models are neural-network-based; ii) even if the metrics are rule-based, the leading distortion might over-adapt to certain assessment models. To avoid these problems, we take a different way of designing an approximated generation loss $\mathcal{L}_{\text{gen}}(X_p; \theta) \in \mathbb{R}^+$
 used for crafting poison. Our design of $\mathcal{L}_{\text{gen}}$ is motivated by the insight that introducing some ``meaningless'' and ``hard-to-understand'' patterns can trick the diffusion model into overfitting on the perturbations, leading the diffusion model to establish false correlations. 
 Specifically, our approximated generation loss can be formulated as,

\begin{equation}
 \mathcal{L}_{\text{gen}}(X_p;\theta) = \mathbb{E}_{\bc, \bx^\prime\sim X_p}\left[\mathcal{L}_{\text{denoise}}(\bx^\prime, \bc; \theta)\right].
    \label{eq. gen_loss}
\end{equation}
Our empirical observation indicates that the maximization of this loss can result in chaotic content and scattered texture in the generated images. 
Instead of introducing an additional targeting loss term \citep{liang2023adversarial,liang2023mist}, we found that our simple denoising-maximization loss is more effective against fine-tuning-based diffusion models.
However, perturbations crafted directly with (\ref{eq. gen_loss}) are fragile to minor transformations and ineffective in bypassing (\ref{eq:bilevel.lower}). 
To remedy this, we adopt the {\it expectation over transformation} technique (EOT \citep{athalye2018synthesizing}) into the PGD process. 
Specifically, given $T$ as a distribution over a set of transformations in (\ref{eq:bilevel.lower}), 
we apply EOT on (\ref{eq:unroll-example}) as

\begin{equation}
\scriptsize
X_{p}^{i+1} = \E_{g\sim T}  \left[\Pi_{B_\infty(X_{p}^{0},r)}(X_{p}^{i} + \alpha \mathrm{sign}(\nabla_{X_{p}^{i}} \mathcal{L}_{\text{gen}}(g(X_p^{i}); \hat{\bx}_{\theta^\prime_{i,K}})) \right], 
\label{eq:EOT}
\end{equation}
where $g(X_p)=\{g(x_p): x_p \in X_p \}$ is the transformed image of $X_p$ under the transformation $g$, $\theta^\prime$ is a K-step unrolled model following (\ref{eq:unroll1}), and the expectation is estimated by Monte Carlo sampling with $J$ samples ($J=1$ in our setup). In summary, we present the overall framework in Alg. \ref{algo:metacloak}.

\begin{algorithm}[t]
\caption{
Training robust perturbation with MetaCloak
}
\begin{algorithmic}[1]
\Require
User training image set $X_c$, training iteration $N_{\text{max}}$, the number of surrogate models $M$, PGD radius $r$, PGD step size $\alpha$, the unrolling number $K$, the transformation distribution $T$, the sampling times $J$, learning rate $\beta$.
\Ensure Protected user image set $X_p$
    \State Stagger the $M$ models as a pool $\mathcal{D}_\theta$, training the $m$-th model weights $\theta_m$ up to $\lfloor m N_{\text{max}} / M\rfloor $ steps on $X_c$
    \For{$i$ \textbf{in} $1, \cdots, C$ crafting steps}
    \Comment{Outer-loop}
            \State Sample a batch of surrogate  $\{\theta_m\}_{m=1}^{B}, \theta_m \sim  \mathcal{D}_\theta $ 
        
            \For{$m$ \textbf{in} $1,\cdots, B$} \Comment{Inner-loop }
                \State Copy model weight $\tilde{\theta}= \theta_m$
            
            \For{ $k=1, \ldots, K$ unroll steps }

            \State  $
            \tilde{\theta}=\tilde{\theta}-\beta \nabla_{\tilde{\theta}} \mathcal{L}_{\text {db }}^{\text{rob}}\left(X_p, T ; \tilde{\theta}\right)
            $
            \EndFor
             \For{$j$ \textbf{in} $1,\cdots, J$}
                    \State 
$\mathcal{L}^j_m=\mathcal{L}_{\text{gen}}\left(t_j(X_p); \tilde{\theta}\right)$, $t_j \sim T$ 
            \EndFor
            \State $\theta_m=\theta_m-\alpha \nabla_{\theta_m} \mathcal{L}_{\text {db }}^{\text{rob}}\left(X_p, T; \theta_m\right)$ 
            \EndFor
            \State Average denoise losses $\mathcal{L}_{\mathrm{adv}}=\sum_{m,j} \mathcal{L}^j_m / BJ$
            
            \State Compute $\nabla_{X_p} \mathcal{L}_{\mathrm{adv}}$ 
            \State Update $X_p$ using SGD and project onto $r$ ball
    \EndFor
   
    \State \Return $X_p$
\end{algorithmic}
\label{algo:metacloak}
\end{algorithm}

%% file: sections/Experiments.tex
\section{Experiments}
\label{sec:exp}

\input{tab/main-table-merged.tex}

\subsection{Setup}
\header{Datasets} Our experiments are performed on human subjects using the two face datasets: {Celeba-HQ} \citep{karras2017progressive} and {VGGFace2} \citep{cao2018vggface2} following Anti-DreamBooth \citep{le_etal2023antidreambooth}. {CelebA-HQ} is an enhanced version of the original CelebA dataset consisting of 30,000 celebrity face images.
{VGGFace2} is a comprehensive dataset with over 3.3 million face images from 9,131 unique identities.
Fifty identities are selected from each dataset, and we randomly pick eight images from each individual and split those images into two subsets for image protection and reference. 
More results on non-face data are in the App. \ref{app. non-face}.

\header{Training Settings} 
The Stable Diffusion (SD) v2-1-base \citep{Rombach_2022_CVPR} is used as the model backbone by default. For Dreambooth training, we fine-tune both the text-encoder and U-Net model with a learning rate of $5 \times 10^{-7}$ and batch size of 2 for 1000 iterations in mixed-precision training mode. 
We consider two training settings: \textit{standard training  (Stand. Training)} and \textit{training with data transformations (Trans. Training)}. For the standard training setting, DreamBooth is trained without performing special pre-processing. For the training with data transformations scenario, we consider transformations including Gaussian filtering with a kernel size of 7, horizontal flipping with half probability, center cropping, and image resizing to 512x512. 
For both the settings, we leverage two inferring prompts with 50 inferring steps, ``a photo of sks person" and ``a DSLR portrait of sks person'' during inference to generate \textit{16} images per prompt.

\noindent\textbf{Baselines and Implementation Details.} 
We compare our method with the following adopted state-of-the-art baselines in \cite{liang2023adversarial,le_etal2023antidreambooth,liang2023mist}: i) \textit{ASPL} \citep{le_etal2023antidreambooth} alternatively update the perturbations and surrogate models, where the surrogate models are updated on both poisons and clean data; ii) \textit{E-ASPL} is an extension of ASPL that ensembles multiple types of diffusion models for better transferability; iii) \textit{FSMG} leverages a DreamBooth trained on clean image for crafting adversarial examples; iv) \textit{AdvDM} \citep{liang2023adversarial, liang2023mist} leverages a pre-trained diffusion model for crafting adversarial examples with additional targeting loss for texture distortion; v) \textit{Glaze} \cite{shan2023glaze} minimizes the representation between the perturbed image and a target dummy image with a pre-trianed encoder; vi) \textit{PhotoGuard} \cite{salman2023raising} perturb image to align its denoised image closer to a dummy target with efficient partial back-propagation; Following the common setup \cite{le_etal2023antidreambooth}, we set the noise radius ($\ell_\infty$-norm ball) to 11/255 with a step size of 1/255 and a step number of 6 by default. We set the unrolling number $K=1$, surrogate model number $M=5$, sample batch size $B=1$, and crafting step $C=4000$ with SD v2-1-base as the surrogate model, which takes about 3 GPU hours to train instance-wise noise. 
See App. \ref{app. exp-details} for more details.

\begin{figure*}[t]
    \centering
    \includegraphics[width=\linewidth]{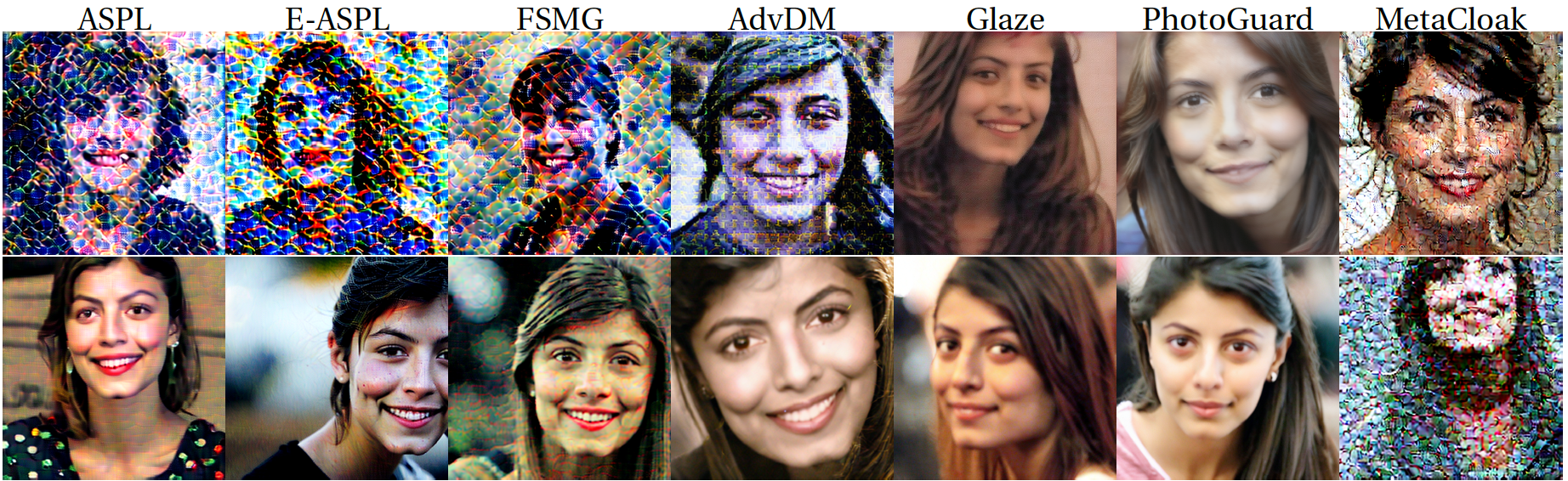}
    \caption{
        Visualization of transformation robustness of different methods. The first row is a generated sample from DreamBooth trained on poisons with no transformation defenses. The 2-th row showcases the robustness of each method under transformation with a Gaussian kernel size of 7. Our method performs robustly under transformation defenses, while other methods fail to preserve the perturbation.
    }
    \label{fig:main.compare.vis}
\vspace{-10pt}
\end{figure*}
\noindent\textbf{Metrics. }\label{sec:standard}
We evaluate the generated images in terms of their semantic-related quality and graphical quality. For the semantic-related score, first, we want to determine whether the subject is present in the generated image. We term this score as \textit{Subject Detection Score (SDS)}. For human faces, we can take the mean of face detection confidence probability using RetinaFace detector \citep{deng2020retinaface} as its SDS. 
Secondly, we are interested in how the generated image is semantically close to its subject. We term this score as \textit{Identity Matching Score (IMS)} \citep{le_etal2023antidreambooth}, the similarity between embedding of generated face images and an average of all reference images. 
We use VGG-Face \cite{serengil2021lightface} and CLIP-ViT-base-32 \cite{radford2021learning} as embedding extractors and employ the cosine similarity. Besides, we use \textit{LIQE} \citep{zhang2023blind}, a vision-language multitask-learning image quality assessment model, for human scene category prediction.
For the graphical quality, we design \textit{CLIP-IQAC}, which is based on CLIP-IQA \cite{wang2022exploring} by considering additional class information.
See the App. \ref{app. metrics} for details. 

\subsection{Effectiveness of MetaCloak}
\label{sec:effectiveness}

\header{Effectiveness Comparison through Quantitative and Qualitative Metrics}
As observed in Tab.~\ref{tab:merged.main.table}, MetaCloak consistently outperforms other baselines across all the metrics. Specifically, in the most important metric, SDS, which measures whether a face appeared in the generated image, MetaCloak successfully degraded this metric by 36.9\% and 45.0\% compared to previous SoTA on VGGFace2 and CelebA-HQ. Regarding reference-based semantic matching metrics, the results on IMS-VGG and IMS-CLIP also show that our method is more effective than other baselines. Regarding image quality metrics, the results on CLIP-IQAC suggest that MetaCloak can effectively degrade the image quality of generated images. Furthermore, the results in Tab. \ref{tab:main-standard-train} (see App.~\ref{app. stand-training}) also suggest that MetaCloak also achieves better effectiveness than existing methods under Stand. Training settings. 
For visualization, as we can see in Fig.~\ref{fig:main.compare.vis}, compared to other baselines, MetaCloak can robustly fool the model to generate images with low quality and semantic distortion under both Stand. and Trans. training while others are sensitive to transformations. More visualizations are in App \ref{app. more-vis-comp}.

\begin{figure}[t]
        \centering
        \includegraphics[width=0.9\linewidth]{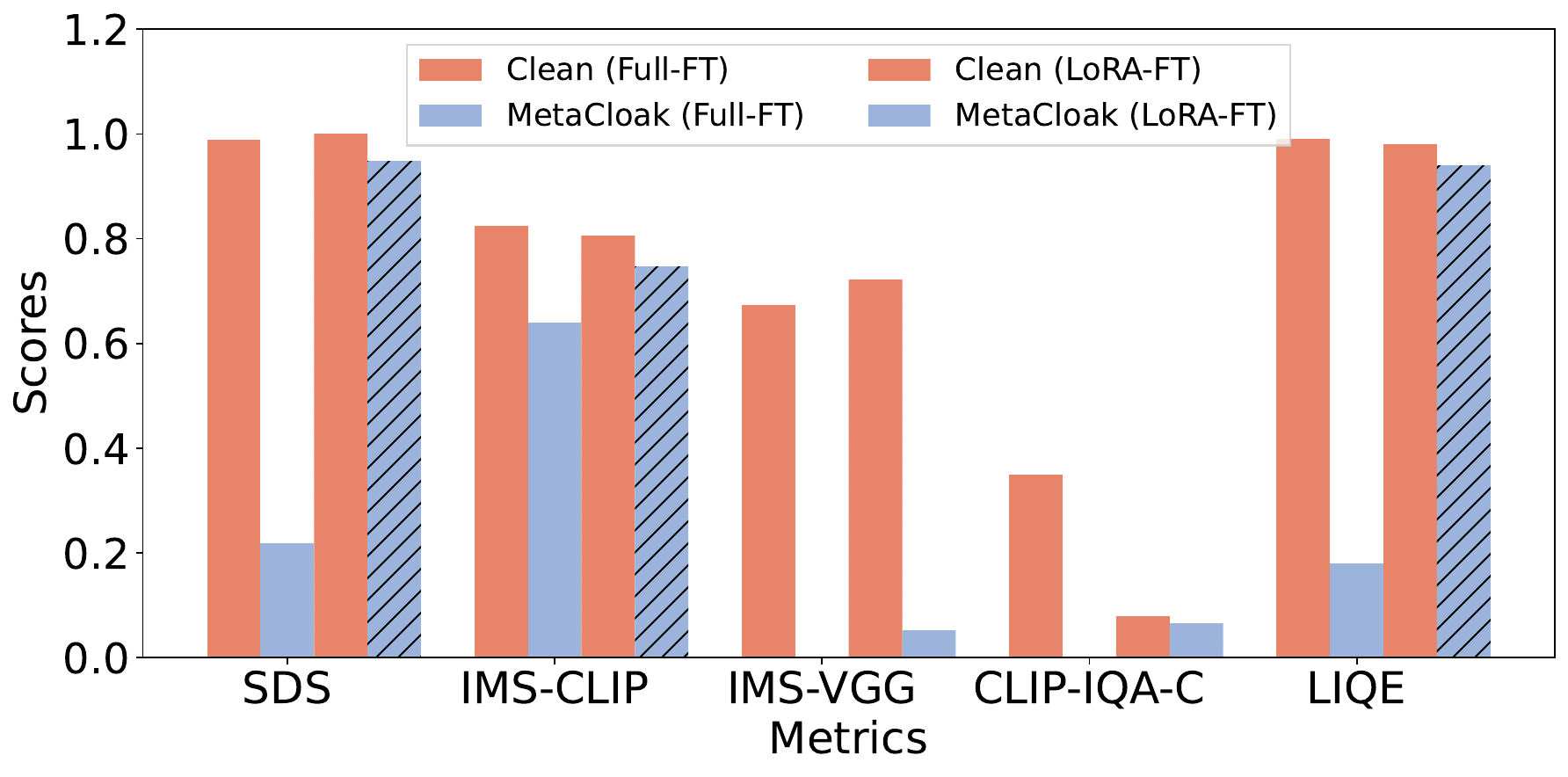}
        \label{fig:full_ft}
    \caption{Results 
    under online training-as-service settings with the Full and LoRA DreamBooth fine-tuning settings on the Replicate. }
    \label{fig:training-as-service}
\end{figure}

\header{Effectiveness under Online Training-as-services Scenarios} To test the effectiveness of our framework in the wild, we conduct experiments under online training-as-service settings. Unlike local training, attacking online training services is more challenging due to the limited knowledge of data prepossessing. 
We first showcase the performance of our method under two common DreamBooth fine-tuning scenarios, including full fine-tuning (Full-FT) and LoRA-fine-tuning (LoRA-FT). We sample data from VGGFace2 and upload its clean and poisoned images to \citet{replicate} for DreamBooth training. 
From the results in Fig. \ref{fig:training-as-service}, we can see that MetaCloak performs well under the Full-FT setting; for instance, it successfully degrades the SDS from 98.9\% to 21.8\%. Under the LoRA-FT setting, the model seems to be more resilient, but MetaCloak can still cause some degradation. {On the inherent defense ability of LoRA-FT, we conjecture that it only fine-tunes a few additional layers of parameters, which might be less likely to overfit and thus more robust to MetaCloak}. 
Additionally, under another popular training setting with the Text Inversion, we validate that MetaCloak can still work well in disturbing the generation quality in Tab. \ref{tab:textInversion}. 
These results demonstrate that MetaCloak can seriously threaten online training services.

\begin{table}[t]
    \includegraphics[width=\linewidth]{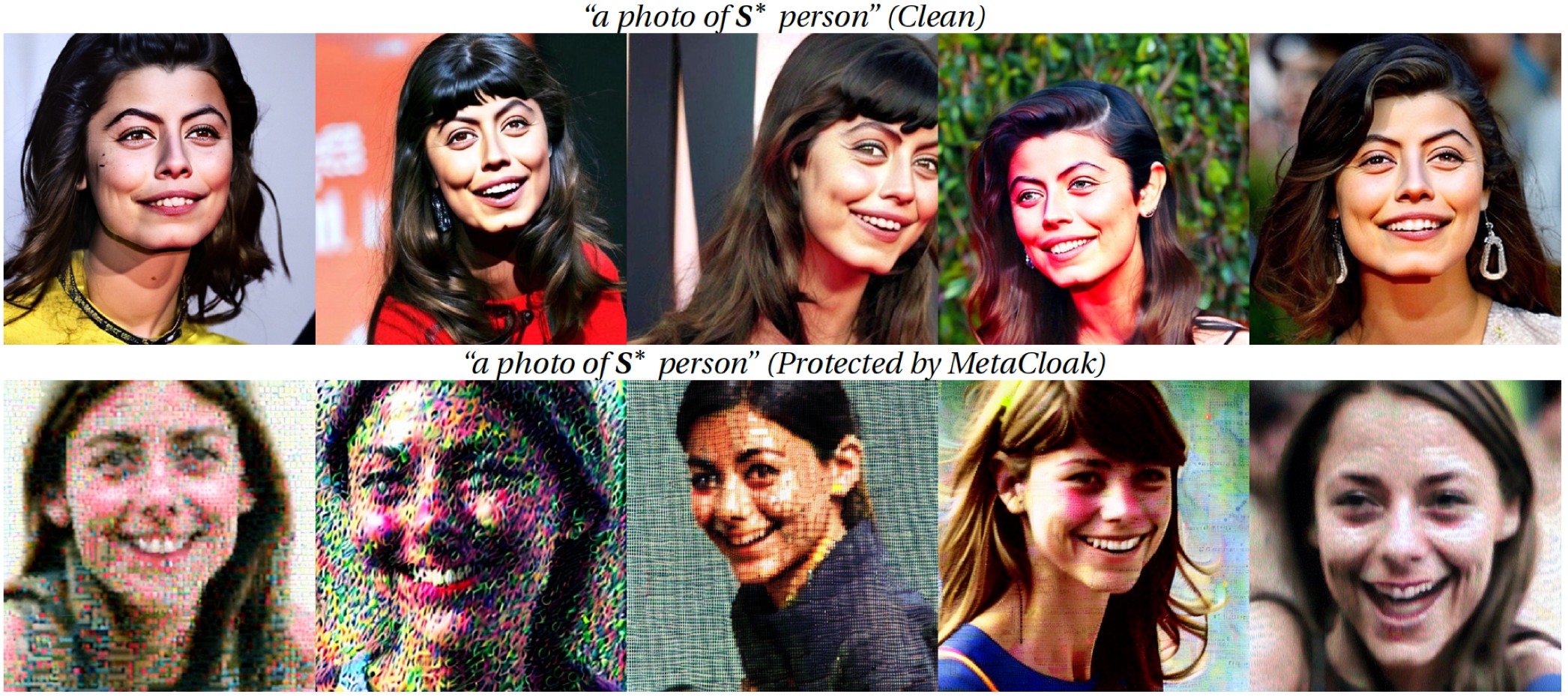}
    \caption{Generated images of Stable Diffusion fine-tuned with Textual Inversion on a randomly selected instance from VGGFace2. $\boldsymbol{S}^*$ denotes a learned token initialized from token \textit{<person>}. }
    \label{tab:textInversion}
\end{table}

\header{Effectiveness of Proposed Components}
To evaluate the individual contributions of MetaCloak's components to its overall effectiveness, we conducted ablation studies using the VGGFace2 dataset, specifically under the Trans. Training setting. The results, presented in Tab. \ref{tab:ablation}, indicate that each module within MetaCloak independently plays a role in degrading the generative performance. Meanwhile, the integration of all modules results in the most robust protection.

\begin{figure}[t]
    \includegraphics[width=\linewidth]{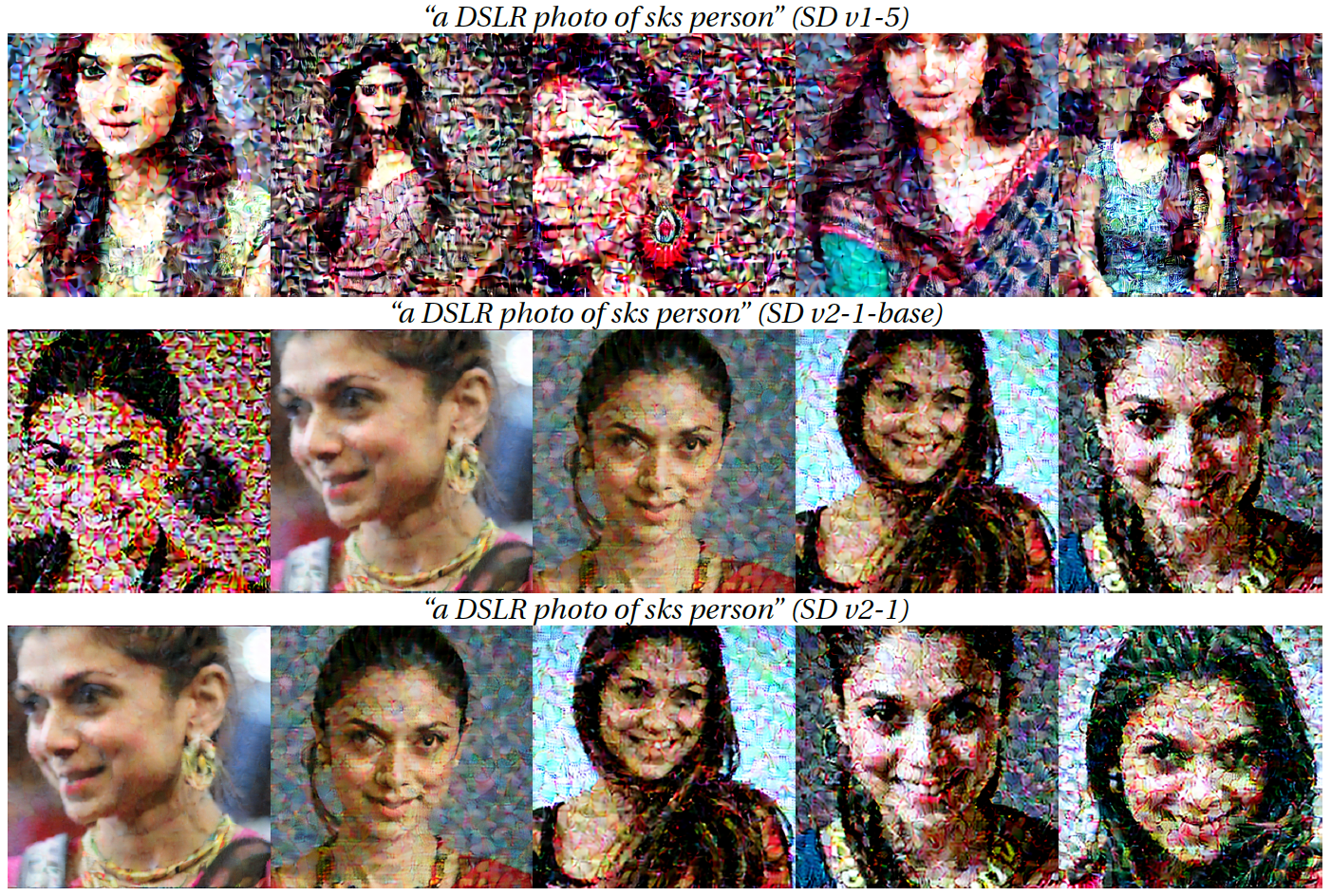}
    \caption{
    Visualization of the generated image of Dreambooth trained on a protected instance from Celeba-HQ with different diffusion models, including SD v1-5, SD v2-1-base, and SD v2-1.
    }
    \label{fig: vis-diff-models-main}
\end{figure}

\begin{figure}[t]
    \centering
    \includegraphics[width=\linewidth]{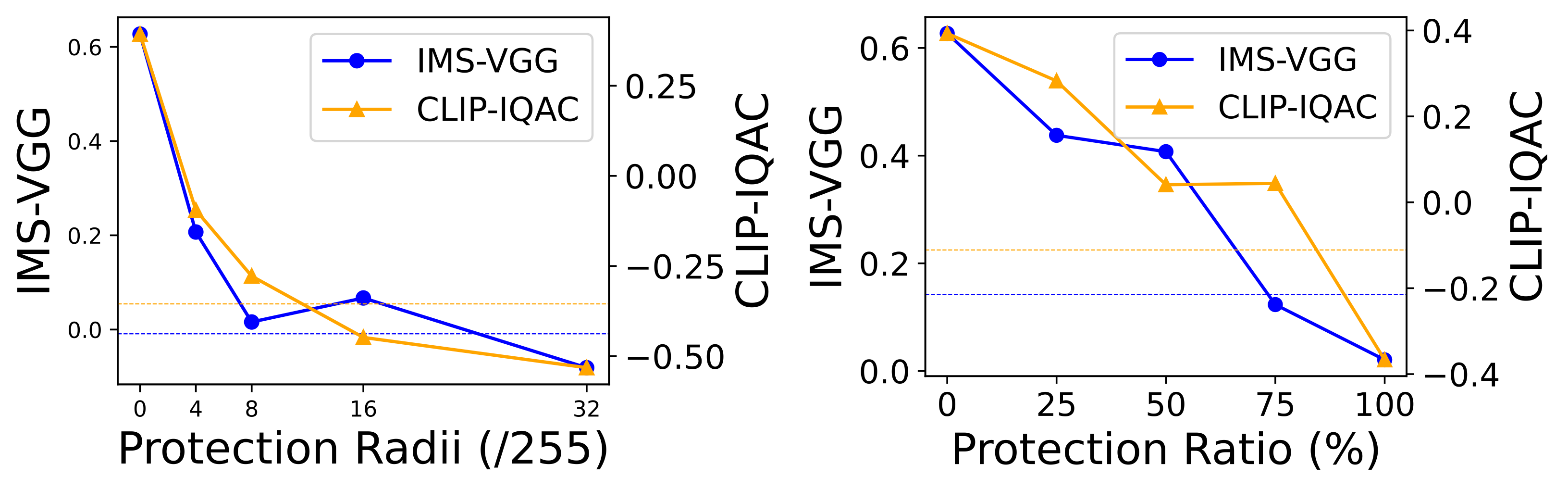}
    \caption{Results of MetaCloak with different protection radii and ratios. The dashed line marks the previous SoTA results under $r=32/255$ and the protection ratio of 100\%, respectively. }
    \label{fig:diff-R-r}
    \vspace{-15pt}
\end{figure}

\header{Effectiveness across Different Training Settings} While crafting specifically with the SD v2-1-base model, one might wonder whether our protection can transfer across training settings with other diffusion models. To investigate that, we further conducted experiments under training settings with two other versions of stable diffusion models, including SD v2-1 and SD v1-5. The results in Fig. \ref{fig: vis-diff-models-main} show that our protection is well transferable in introducing distortion in generation across different versions of stable diffusion models. Furthermore, we also study training settings with different protection ratios and radii under the Trans. Training settings on VGGFace2 dataset. As shown in Fig.~\ref{fig:diff-R-r}, these two factors are essential for protection. Besides, MetaCloak is more budget-efficient since it matches the performance of the previous SoTA with lower protection ratios and radii. Please refer to the App. \ref{app.radii} on the trade-off between stealthiness and performance of our perturbation under different radii. 

\begin{table}[thbp]
    \centering
\resizebox{\linewidth}{!}{
\begin{tabular}{ccccccc} 
            \toprule
           Meta. &EOT   & SDS                        & $IMS_{\text{CLIP}}$          & $IMS_{\text{VGG}}$       & CLIP-IQAC                    & LIQE        \\
            \midrule
         
      $\times$   &   $\times$  & 0.879  & 0.712   & 0.193  & -0.090   & 0.891 \\
            
   $\checkmark$ &   $\times$    & 0.787  & 0.692  & 0.056  & -0.108  & 0.871  \\
        
   $\times$ & $\checkmark$   & 0.515 & 0.677 & 0.072 & -0.244  & 0.570  \\

           $\checkmark$ & $\checkmark$&  \textbf{0.432}  & \textbf{0.644} & \textbf{-0.151} & \textbf{-0.440} & \textbf{0.562}\\
        \bottomrule 
\end{tabular}

}
    \caption{
    Ablation Study of MetaCloak on VGGFace2 under Trans. Training. The 2nd to 4th rows are the ablated versions. ``Meta.'' denotes the meta-learning process in our method. 
    }\label{tab:ablation}
\end{table}

\begin{table}[t]
    \centering
    \resizebox{\linewidth}{!}{
\begin{tabular}{ccccccc} 
\toprule
Setting & Defenses & 
SDS                        & $IMS_{\text{CLIP}}$         & $IMS_{\text{VGG}}$         & CLIP-IQAC                    & LIQE        \\
\midrule
\multirow{5}{*}{
   \begin{tabular}[c]{@{}c@{}}Stand.\\Training\end{tabular}} & $\times$ & 0.296  & 0.662   & -0.051   & -0.380  & 0.180 \\
 & +SR & \textbf{0.876}   &\textbf{ 0.748}   & \textbf{0.354} & \textbf{0.417}  & \textbf{0.984} \\
 & +TVM & 0.690  & 0.638 & 0.320 & -0.083  & 0.867 \\

& +JPEG & 0.496  & 0.682  & -0.135  & -0.363  & 0.365 \\
\cmidrule(lr){2-7}
& Oracle* & 0.897 & 0.814   & 0.438  & 0.456   & 0.992  \\
\midrule
\multirow{5}{*}{   \begin{tabular}[c]{@{}c@{}}Trans.\\Training\end{tabular}}   & $\times$ & 0.432 & 0.644 & -0.151 & -0.440 & 0.570 \\
& +SR   &   \textbf{0.824}  & \textbf{0.739}  & \textbf{0.507}  & \textbf{0.039}  & \textbf{0.945} \\
& +TVM  & 0.617  & 0.636   & 0.175   & -0.076 & 0.771  \\
& +JPEG  & 0.616  & 0.691  & -0.160  & -0.262  & 0.562 \\
\cmidrule(lr){2-7}
& Oracle* & 0.903  & 0.790  & 0.435   & 0.329  & 0.984  \\
\bottomrule
\end{tabular}
    }
    \caption{
    Resilience of MetaCloak under more advanced adversarial purifications. JPEG compression, Super-resolution (SR), and Total-variation minimization (TVM) are considered. Oracle* denotes the performance of Dreambooth trained on clean data.
}
    \label{tab: more-defenses.main.table}
\end{table}

\subsection{Resistance against Adversarial Purification}

We consider three adversarial purification techniques, including JPEG compression \citep{liu2019feature}, super-resolution (SR) \citep{mustafa2019image}, and image reconstruction based on total-variation minimization (TVM) \citep{wang2020adversarial}. We use a quality factor of 75 for the JPEG defense and a scale factor of 4 for the SR defense \citep{liang2023adversarial}.   
As shown in Tab.~\ref{tab: more-defenses.main.table}, additional purification does hinder the data protection performance of MetaCloak to some extent, and SR is the most effective one in restoring the generation ability compared to others. 
For the other two approaches, we found that JPEG compression will compromise the purified image quality, and the TVM does better in retaining graphical quality, but it causes significant and unacceptable distortion in the face region of purified images. 
While SR does the best in purification, it still can not fully recover the original generation ability of DreamBooth trained on clean data. 
See App. \ref{app. adv.puri} for more details and results. 

%% file: tab/main-table-merged.tex
\begin{table*}[thbp]
    \centering
    \resizebox{.95\linewidth}{!}{
        \begin{tabular}{ccccccc} 
            \toprule
              Dataset            & Method    & SDS $\downarrow$                       & $IMS_{\text{CLIP}}$    $\downarrow$      & $IMS_{\text{VGG}}$  $\downarrow$     & CLIP-IQAC $\downarrow$                   & LIQE $\downarrow$       \\
            \midrule
             \multirow{7}{*}{VGGFace2} &  Clean   
            & 0.903 {\scriptsize{$\pm$ 0.291}}  & 0.790 {\scriptsize{$\pm$ 0.076}}  & 0.435 {\scriptsize{$\pm$ 0.657}}  & 0.329 {\scriptsize{$\pm$ 0.354}}  & 0.984 {\scriptsize{$\pm$ 0.124}} \\

                                                                  & ASPL     
                                                                & 0.879 {\scriptsize{$\pm$ 0.321}}  & 0.712 {\scriptsize{$\pm$ 0.082}}  & 0.193 {\scriptsize{$\pm$ 0.792}}  & -0.090 {\scriptsize{$\pm$ 0.432}}  & 0.891 {\scriptsize{$\pm$ 0.312}} \\

                                                                  & EASPL    
                                                                & \secondrunner{0.801} {\scriptsize{$\pm$ 0.395}}  & \secondrunner{0.703} {\scriptsize{$\pm$ 0.076}}  & \secondrunner{0.142} {\scriptsize{$\pm$ 0.814}}  & -0.024 {\scriptsize{$\pm$ 0.406}}  & \secondrunner{0.820} {\scriptsize{$\pm$ 0.384}} \\

                                                                  & FSMG     
                                                                & 0.911 {\scriptsize{$\pm$ 0.280}}  & 0.718 {\scriptsize{$\pm$ 0.063}}  & 0.273 {\scriptsize{$\pm$ 0.753}}  & \secondrunner{-0.111} {\scriptsize{$\pm$ 0.371}}  & 0.891 {\scriptsize{$\pm$ 0.312}} \\
                                                               & AdvDM    
                                                            & 0.903 {\scriptsize{$\pm$ 0.291}}  & 0.769 {\scriptsize{$\pm$ 0.068}}  & 0.570 {\scriptsize{$\pm$ 0.510}}  & 0.241 {\scriptsize{$\pm$ 0.341}}  & 0.984 {\scriptsize{$\pm$ 0.124}} \\
                                                            & Glaze     & 0.910 {\scriptsize{$\pm$ 0.279}}  & 0.774 {\scriptsize{$\pm$ 0.083}}  & 0.490 {\scriptsize{$\pm$ 0.607}}  & 0.258 {\scriptsize{$\pm$ 0.330}}  & 0.984 {\scriptsize{$\pm$ 0.124}} \\
                                                                    & PhotoGuard     
                                                         & 0.928 {\scriptsize{$\pm$ 0.255}}  & 0.793 {\scriptsize{$\pm$ 0.072}}  & 0.524 {\scriptsize{$\pm$ 0.558}}  & 0.407 {\scriptsize{$\pm$ 0.249}}  & 1.000 {\scriptsize{$\pm$ 0.000}} \\

                                                                  & MetaCloak 
                                                                & \textbf{0.432}* {\scriptsize{$\pm$ 0.489}}  & \textbf{0.644}* {\scriptsize{$\pm$ 0.107}}  & \textbf{-0.151}* {\scriptsize{$\pm$ 0.864}}  & \textbf{-0.440}* {\scriptsize{$\pm$ 0.223}}  & \textbf{0.570}* {\scriptsize{$\pm$ 0.495}} \\
 
                \midrule
              \multirow{7}{*}{CelebA-HQ} & Clean      & 0.810 {\scriptsize{$\pm$ 0.389}}  & 0.763 {\scriptsize{$\pm$ 0.119}}  & 0.181 {\scriptsize{$\pm$ 0.784}}  & 0.470 {\scriptsize{$\pm$ 0.264}}  & 0.984 {\scriptsize{$\pm$ 0.124}} \\

                                                                  & ASPL      
                                                                & \secondrunner{0.755 } {\scriptsize{$\pm$ 0.427}}  & 0.714 {\scriptsize{$\pm$ 0.082}}  & 0.044 {\scriptsize{$\pm$ 0.823}}  & 0.054 {\scriptsize{$\pm$ 0.364}}  & 0.969 {\scriptsize{$\pm$ 0.174}} \\

                                                                 & EASPL     
                                                                & 0.841 {\scriptsize{$\pm$ 0.362}}  & \secondrunner{0.707 }{\scriptsize{$\pm$ 0.069}}  & \secondrunner{-0.113} {\scriptsize{$\pm$ 0.834}}  & \secondrunner{0.020} {\scriptsize{$\pm$ 0.423}}  & 0.984 {\scriptsize{$\pm$ 0.124}} \\

                                                                  & FSMG     
                                                                & 0.769 {\scriptsize{$\pm$ 0.416}}  & 0.718 {\scriptsize{$\pm$ 0.081}}  & 0.073 {\scriptsize{$\pm$ 0.805}}  & 0.085 {\scriptsize{$\pm$ 0.343}}  & \secondrunner{0.953} {\scriptsize{$\pm$ 0.211}} \\

                                                                  & AdvDM     
                                                                  & 0.866 {\scriptsize{$\pm$ 0.339}}  & 0.789 {\scriptsize{$\pm$ 0.083}}  & 0.431 {\scriptsize{$\pm$ 0.618}}  & 0.373 {\scriptsize{$\pm$ 0.268}}  & 1.000 {\scriptsize{$\pm$ 0.000}} \\

                                                                  & Glaze     
                                                                  & 0.889 {\scriptsize{$\pm$ 0.312}}  & 0.778 {\scriptsize{$\pm$ 0.092}}  & 0.318 {\scriptsize{$\pm$ 0.716}}  & 0.338 {\scriptsize{$\pm$ 0.311}}  & 0.992 {\scriptsize{$\pm$ 0.088}} \\

                                                                    & PhotoGuard     
                                                           & 0.879 {\scriptsize{$\pm$ 0.320}}  & 0.790 {\scriptsize{$\pm$ 0.092}}  & 0.401 {\scriptsize{$\pm$ 0.658}}  & 0.451 {\scriptsize{$\pm$ 0.277}}  & 0.992 {\scriptsize{$\pm$ 0.088}} \\

                                                                  & MetaCloak   &
                                                                  \textbf{0.305}* {\scriptsize{$\pm$ 0.452}}  & \textbf{0.608}* {\scriptsize{$\pm$ 0.109}}  & \textbf{-0.637}* {\scriptsize{$\pm$ 0.686}}  & \textbf{-0.354}*{\scriptsize{$\pm$ 0.244}}  & \textbf{0.438}* {\scriptsize{$\pm$ 0.496}} \\

            \bottomrule
        \end{tabular}
    }
    \caption{
        Results of different methods under the transformations training setting with the corresponding mean and standard deviation (±) on two datasets. The best data performances are in \textbf{bold}, and the second runners are shaded in gray. The Gaussian filtering kernel size is set to 7. * denotes a significant improvement according to the Wilcoxon signed-rank significance test ($p \leq 0.01$). 
    }
    \vspace{-8pt}
    \label{tab:merged.main.table}
\end{table*}

%% file: sections/conclusion.tex
\section{Conlusion}
\label{sec:conclusion}
This paper proposes MetaCloak as the first work that protects user images from unauthorized subject-driven text-to-image synthesis under data transformation purification. We resolve the limitations of existing works in sub-optimal optimization and fragility to data transformations with a novel meta-learning framework and transformation-robust perturbation crafting process. Extensive experiments demonstrate the effectiveness of MetaCloak. Future directions include establishing mechanism interpretations and designing more efficient perturbations.
\header{Acknowledgement} 
In this work, Yixin Liu, Yutong Dai, and Lichao Sun were supported by the National Science Foundation Grants CRII-2246067.

%% file: sections/appendix/details.tex
\section{Experiment Details}
\label{app. exp-details}
\subsection{Hardware and DreamBooth Training Details}
All the experiments are conducted on an Ubuntu 20.04.6 LTS (focal) environment with 503GB RAM, 10 GPUs (NVIDIA\textsuperscript{\textregistered} RTX\textsuperscript{\textregistered} A5000 24GB), and 32 CPU cores (Intel\textsuperscript{\textregistered} Xeon\textsuperscript{\textregistered} Silver 4314 CPU @ 2.40GHz). Python 3.10.12 and Pytorch 1.13 are used for all the implementations. For the DreamBooth full trianing mode, we use the 8-bit Adam optimizer \citep{kingma2017adam} with $\beta_1 = 0.9$ and $\beta_2 = 0.999$ under bfloat16-mixed precision and enable the xformers for memory-efficient training. For calculating prior loss, we use 200 images generated from Stable Diffusion v2-1-base with the class prompt ``a photo of a person''. The weight for prior loss is set to 1. For the instance prompt, we use ``a photo of sks person''. During the meta-learning process, we regularly delete the temporary models and store the surrogates back to the CPU to save GPU memory. It takes about 3 GPU hours to craft perturbations for an instance under this strategy.  

\section{Baseline Methods and Metrics}

\subsection{Evaluation Metrics}
\label{app. metrics}
In this section, we describe the evaluation metrics used in our experiments in more detail. For our proposed CLIP-IQAC, we calculate the CLIP score difference between ``a good photo of [class]'' and ``a bad photo of [class]''. For calculating SDS and IMS-VGGNet, we leverage the APIs for face recognition and face embedding extraction in the deep face library \cite{serengil2021lightface}. In terms of graphical quality, we found that the commonly used metric, BRISQUE \cite{Brisque} is not a faithful metric when we conduct additional data transformations like Gaussian Filtering. We thus omit this score when presenting results in the main text. For instance, the BRISQUE score of a fully poisoned Dreambooth is better than the clean one, as shown in Tab. \ref{tab: low-poisoning-rate.main.table}. Among all the metrics considered, we found that SDS and IMS-VGG are more aligned with our perception of evaluating Dreambooth's personalized generation performance. The SDS score indicates whether a subject is presented in the generated image, while the IMS-VGGNet score measures the similarity between the generated image and the subject. Compared to graphical distortion, semantic distortion is more important when the user wants to prevent the unauthorized generation of their images. Overall, MetaCloak achieves the best performance among all the considered baselines. 

\noindent \textbf{Migrating the Metrics Variance}. To migrate the variance for calculating metrics, we obtain the mean and standard variance in the following manner. For the instance $i$ and its $j$-th metric, its $k$-th observation value is defined as $m_{i,j,k}$. For the $j$-th metric, the mean value is obtained with $\sum_{i,k}m_{i,j,k}/(N_i N_k)$, where $N_i$ is the instance number for that particular dataset, and $N_k$ is the image generation number. And the standard variance is obtained with the formula $\sqrt{\sum_{i,k}(m_{i,j,k} - \mu_j)^2/(N_i N_k)}$, where $\mu_j$ is the mean of the $j$-th metric. Moreover, the Wilcoxon signed-rank significance test for the $j$-th metric between methods is conducted on sequence $\{m_{i,j,k}\}_{i,k}$. We found that these measures provide more faithful indicators of quality change compared to statistics solely over instances population. 

%% file: sections/appendix/more-res.tex
\section{More Experiments Results}
\label{app. more-exp}
\subsection{Results under Standard Training }
\label{app. stand-training}
\input{tab/main-std-train}
We present the results of MetaCloak under the standard training setting in Tab. \ref{tab:main-standard-train}. As we can see, MetaCloak can effectively degrade the DreamBooth's personalized generation performance under the standard training setting. In terms of reference-based semantic metrics, we can see that our method is better in data protection than the previous SOTA in terms of IMS-VGG. Since VGGFaceNet is specially trained on facial datasets and thus more aligned with facial representation, we believe the results on IMS-VGG are more convincing than those on IMS-CLIP. In terms of other metrics, our method is also more effective than the previous SoTA.

\subsection{More Visualizations}
\label{app. more-vis-comp}
\begin{figure*}
    \centering
    \includegraphics[width=\linewidth]{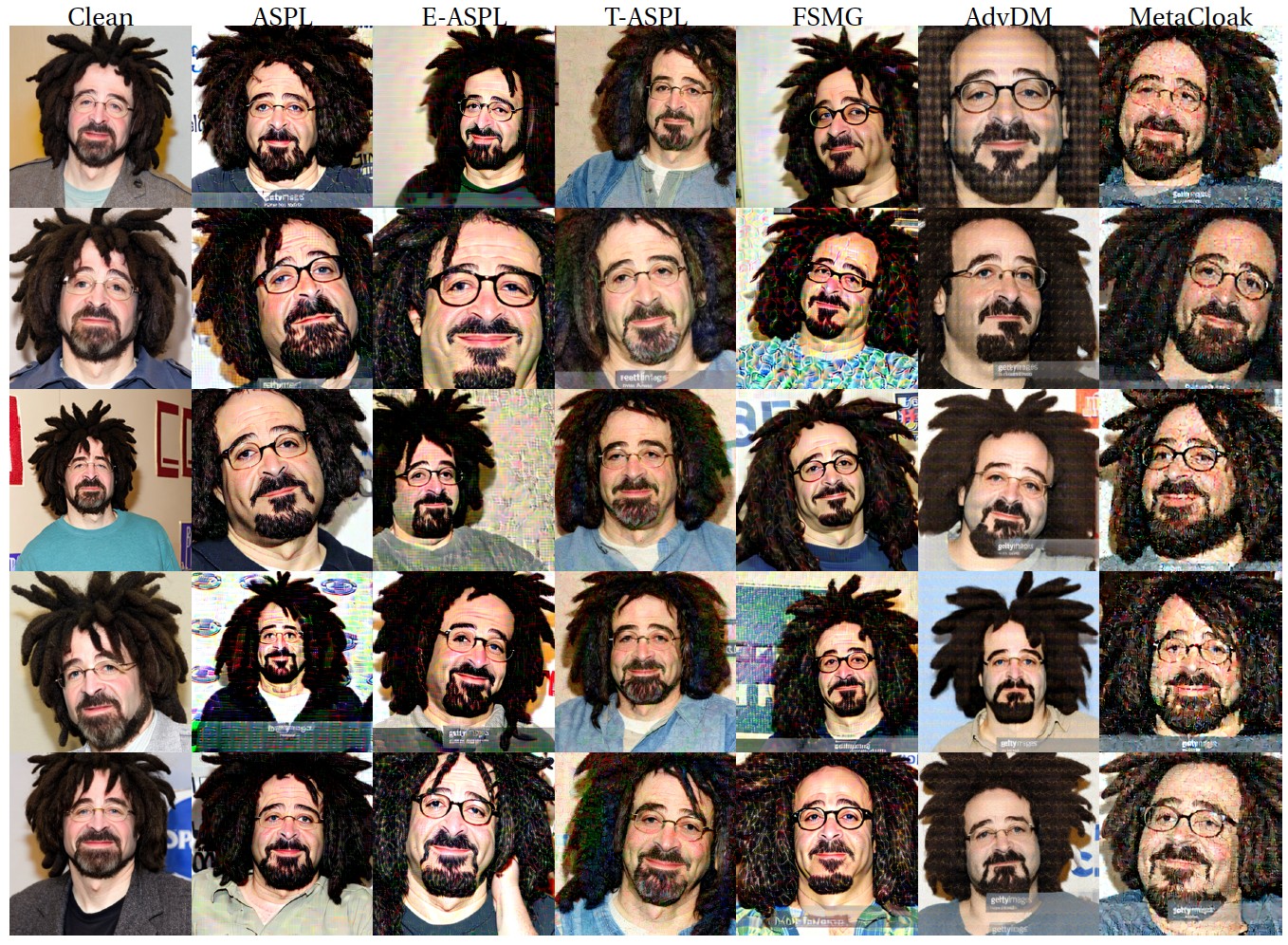}
    \caption{
        Visualization of Dreambooth's generated images on VGGFace2. DreamBooths are trained on data perturbed by different methods under Trans. Training. The first column denotes the Dreambooth trained on clean data. The inferring prompt is ``\promptOne''. 
    }
    \label{fig: vis-comp-methods-2}
\end{figure*}

We visualize the generated images of Dreambooth trained on data perturbed by MetaCloak and other baselines in Fig.~\ref{fig: vis-comp-methods-2}. As we can see, compared to other baselines, MetaCloak can robustly fool DreamBooth into generating images with low quality and semantic distortion under data transformations. In contrast, other baselines are sensitive to data transformation defenses. In this setting, DreamBooth's generation ability is retained since the images generated are of high quality. These results demonstrate that MetaCloak is more robust in defense of data transformation.

\subsection{More Datasets}
\label{app. non-face}
To validate the effectiveness of MetaCloak on non-face data, we conduct additional experiments on the DreamBooth-official subject dataset \cite{ruiz2022dreambooth}, which includes 30 subjects of 15 different classes. Nine out of these subjects are live subjects (dogs and cats), and 21 are objects. Two inference prompts are randomly selected for quality evaluation. As shown in Tab.~\ref{tab:non-face-result}, the results demonstrate that our method can still successfully degrade the generation performance on those inanimate subjects. 

\begin{table*}[thbp]
   \includegraphics[width=\linewidth]{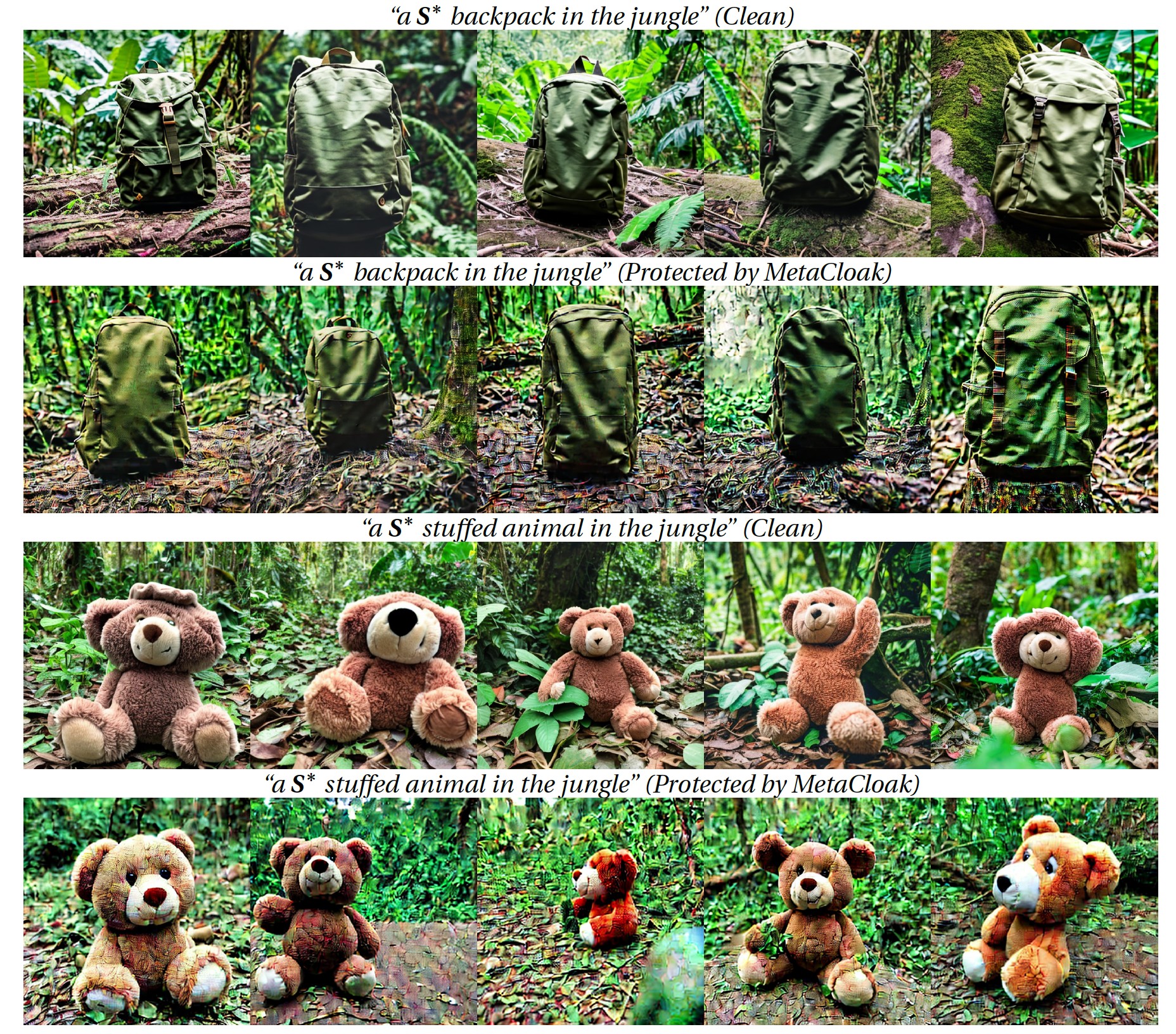}
    \caption{More results of MetaCloak on the Dreambooth-official dataset under the Trans. Training setting. The identifier $\boldsymbol{S}^*$ is set to \textit{sks}. }
    \label{tab:non-face-result}
\vspace{-10pt}
\end{table*}

\subsection{Training DreamBooth on Replicate}
\begin{figure*}[htbp]
    \centering
    \includegraphics[width=\linewidth]{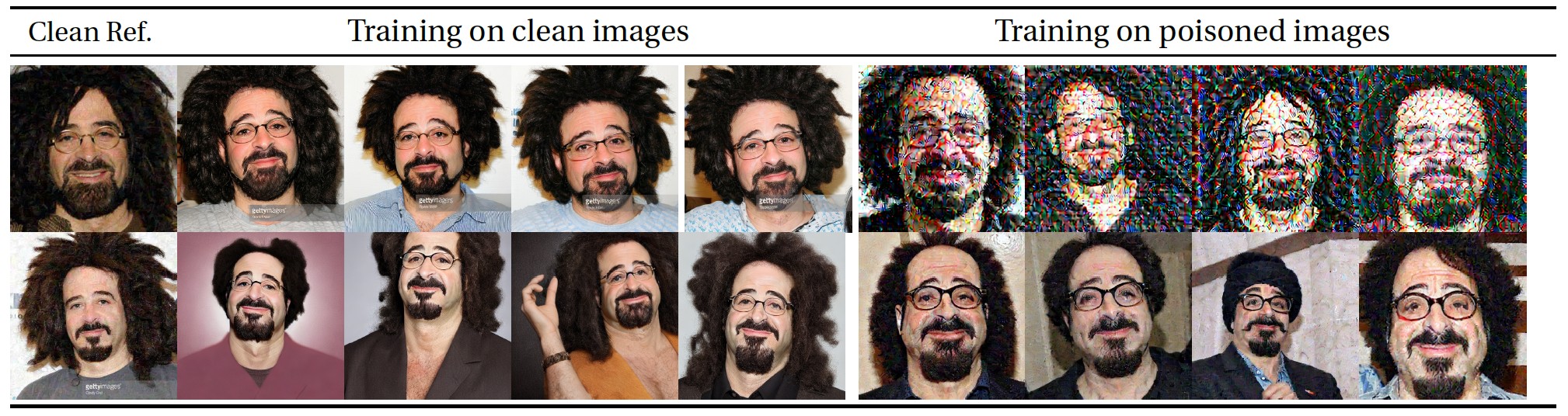}
    \caption{Effectiveness of our method in the wild. Dreambooth training on the replicate platform under two training settings, including full fine-tuning and LoRA-based fine-tuning.}
    \label{fig: training-as-service}
  \end{figure*}

We test the effectiveness of MetaCloak in the wild by training DreamBooth on the Replicate platform \citep{replicate}. The Replicate platform is an online training-as-service platform that allows users to upload their own images and train DreamBooth on them. The generated image of the trained DreamBooth is shown in Fig. \ref{fig: training-as-service}. As we can see, MetaCloak can effectively degrade DreamBooth's personalized generation performance in this setting. As can be seen, MetaCloak can effectively degrade the personalized generation performance of DreamBooth under both Full-FT and LoRA-FT settings, demonstrating that MetaCloak can seriously threaten Dreambooth's online training services.  

\label{app. more-replicate}

\subsection{More Results on Adversarial Purification}
\label{app. adv.puri}

\begin{figure*}
    \centering
    \includegraphics[width=\linewidth]{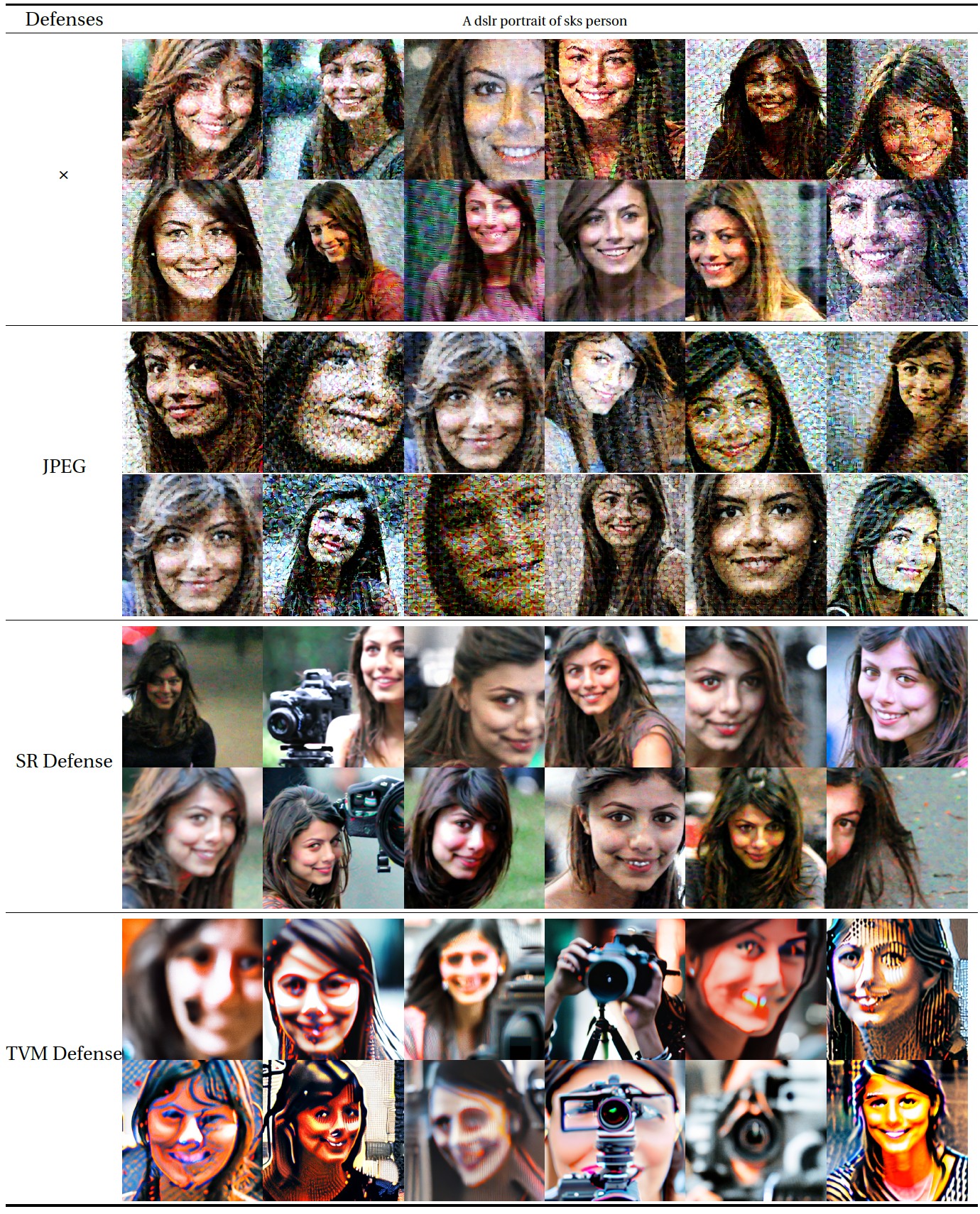}
  \caption{
   Visualizations of generated images of Dreambooth trained with various adversarial purifications under Trans. Training setting.
  }
  \label{fig: visualizations-more-defenses}
\end{figure*}

In the SR defense, we first conduct image resizing with a scale factor of 1/4 and then use the SR model to reconstruct the image. In the TVM defense, we first resize the image to a size of 64x64 for computation feasibility. Then, we use the TVM model to reconstruct the image and then conduct two super-resolution processes and one resize process to align the image size with the original image. The DreamBooth trained on data purificated by JPEG compression, SR, and TVM are shown in Fig. \ref{fig: visualizations-more-defenses}. As we can see, SR defense is the only one that can effectively purify the adversarial perturbation while maintaining the image quality. Compared to SR defense, TVM defense distorts the face significantly, and JPEG defense introduces some artifacts to the image. 

\header{Details of TVM optimization} We implemented the TVM defense in the following steps:
\begin{enumerate}
\item Resize the instance image to $64 \times 64$ pixels. This is done for the computational feasibility of the optimization problem.
\item Generate a random dropout mask $X \in \mathbb{R}^{64 \times 64 \times 3}$ using a Bernoulli distribution with probability $p = 0.02$. This represents the pixel dropout rate.
\item Solve the TVM optimization problem:
\begin{equation}
\min_{Z} ||(1 - X) \odot (Z - x)||_2 + \lambda_{TV}TV_2(Z) ,
\end{equation}
where $x$ is the original image, $Z$ is the reconstructed one, and $\lambda_{TV}$ is the weight on TV term. The first term ensures $Z$ stays close to the original image $x$, and the second term minimizes the total variation, which reduces noise and enforces smoothness.
\item Reshape the optimized $Z$ back to the original size $64 \times 64 \times 3$.
\item Conduct two super-resolution steps (with a resizing process in the middle) to upsample to $512 \times 512$ pixels.
\end{enumerate}

\header{More Results against Advanced Purification} To demonstrate the effectiveness of our method against the latest purification, we additionally present the performance of MetaCloak against two more recent purification methods, DiffPure \cite{nie2022diffusion} and IMPRESS \cite{cao2024impress} in Tab. \ref{tab: more-defenses.main.table}. We set the purification iteration for \texttt{IMPRESS} as $N=3$k, and the optimal perturbation time steps for \texttt{DiffPure} as $t^*=8$. The results show that these purifications still can't fully recover the original generation performance, demonstrating MetaCloak's robustness.

\begin{table}[t]
    \centering
    \resizebox{\linewidth}{!}{
\begin{tabular}{cccccc} 
\toprule
Defenses & 
SDS                        & $IMS_{\text{CLIP}}$         & $IMS_{\text{VGG}}$         & CLIP-IQAC                    & LIQE        \\
\midrule
$\times$ & 0.401 {\scriptsize{$\pm$ 0.485}}  & 0.641 {\scriptsize{$\pm$ 0.118}}  & -0.225 {\scriptsize{$\pm$ 0.852}}  & -0.314 {\scriptsize{$\pm$ 0.266}}  & 0.445 {\scriptsize{$\pm$ 0.497}}  \\

+IMPRESS & 0.541 {\scriptsize{$\pm$ 0.493}}  & 0.691 {\scriptsize{$\pm$ 0.080}}  & -0.074 {\scriptsize{$\pm$ 0.858}}  & -0.307 {\scriptsize{$\pm$ 0.222}}  & 0.789 {\scriptsize{$\pm$ 0.408}} \\ 
+DiffPure  & 0.743 {\scriptsize{$\pm$ 0.430}}  & 0.640 {\scriptsize{$\pm$ 0.081}}  & 0.361 {\scriptsize{$\pm$ 0.673}}  & 0.020 {\scriptsize{$\pm$ 0.289}}  & 0.938 {\scriptsize{$\pm$ 0.242}} \\
%
\midrule
Oracle* & 0.903  & 0.790  & 0.435   & 0.329  & 0.984  \\
\bottomrule
\end{tabular}
    }
    \caption{
    Resilience against advanced defenses on VGGFace2 under \textit{Trans. Training}. Oracle* denotes training on clean data.
}
    \label{tab: more-defenses.main.table}
\end{table}

\subsection{Trade-off of Effectiveness and Stealthiness}
\label{app.radii}
\begin{figure*}[htbp]
    \centering
    \includegraphics[width=\linewidth]{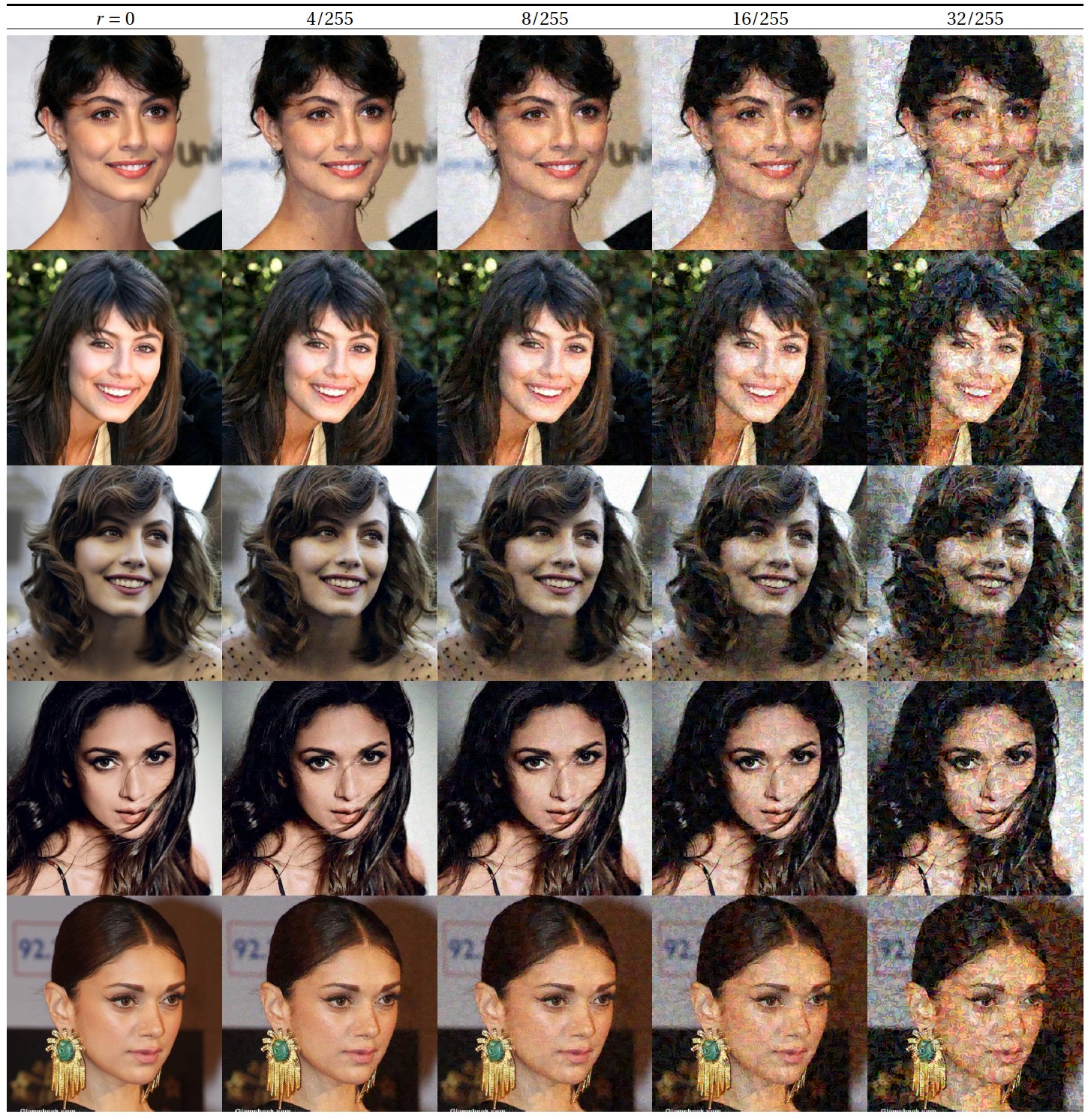}
    \caption{Visualization of perturbed images from VGGFace2 under different attack radii. }
    \label{fig: vis-imgs-diff-radii}
\end{figure*}

To study the effectiveness of MetaCloak under different radii, we conducted experiments with different radii under the Trans. Training setting. As shown in Tab.~\ref{tab:radius.main.table}, increasing the radius can effectively improve the effectiveness of MetaCloak. However, when the radius is too large, the stealthiness of injected noise will also be compromised since some specific noise patterns will overwhelm the image content, as shown in Fig. \ref{fig: vis-imgs-diff-radii}. We conclude that the study of how to further improve the stealthiness of MetaCloak under large radii is an important future direction.

\subsection{Resilience under Low Poisoning Ratio}
\begin{figure*}[htbp]
 \centering
    \includegraphics[width=\linewidth]{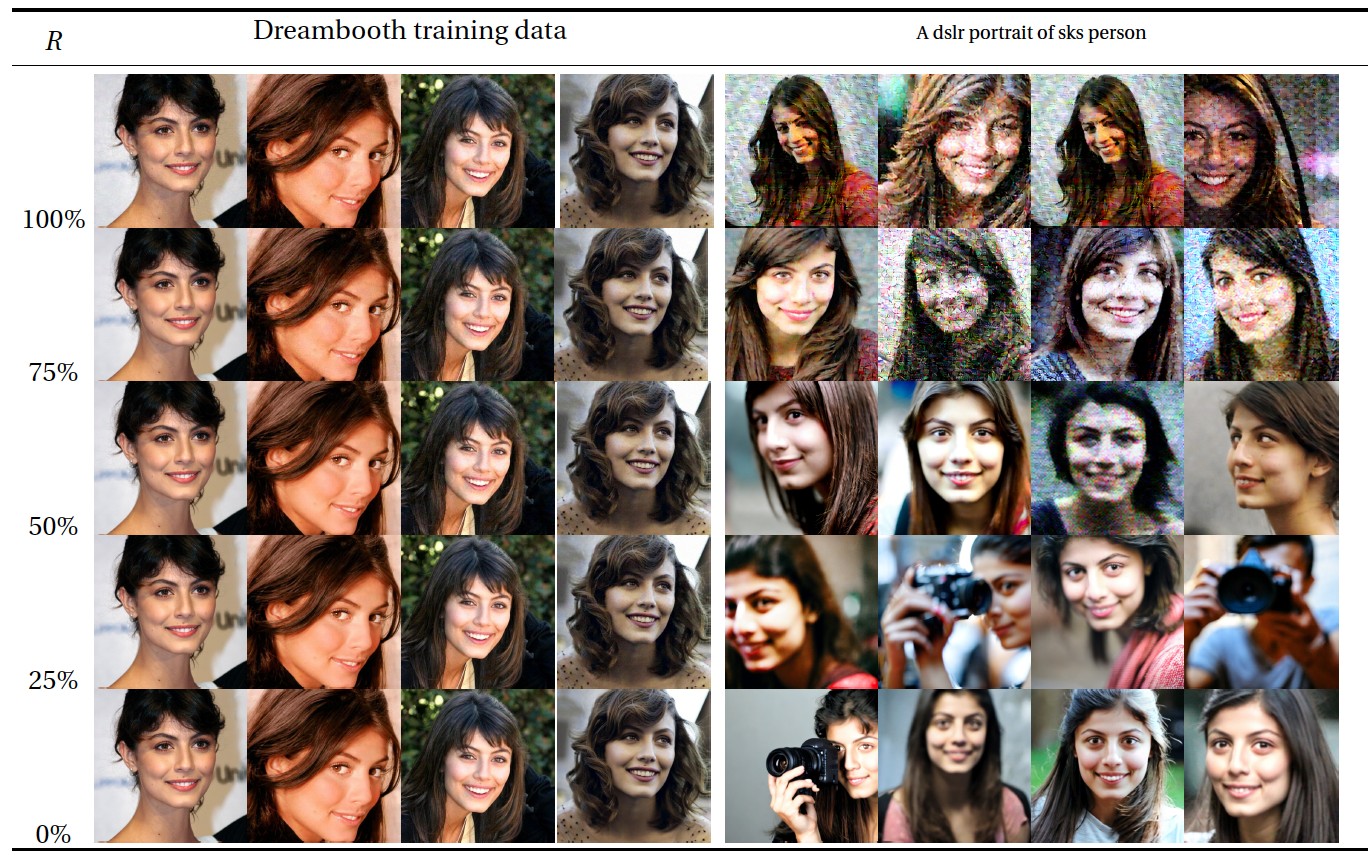}
  \caption{Performance of MetaCloak under different poisoning rates $R$. The Inferring prompt is ``\promptTwo''.}
  \label{fig: diff-rate-1}
\end{figure*}

To study the effectiveness of MetaCloak under low poisoning rates, we conduct experiments with different poisoning rates from $\{0\%, 25\%, 50\%, 75\%, 100\%\}$ under the two training settings. As shown in Tab.~\ref{tab: low-poisoning-rate.main.table} and Fig.~\ref{fig: diff-rate-1}, increasing the poisoning rate can effectively improve the effectiveness of MetaCloak. However, when the poisoning rate is too low, the effectiveness of MetaCloak will be compromised since there is some knowledge leakage. How to effectively protect data under a low poisoning rate is an important future direction. 

\begin{table*}[thbp]
    \centering
    \resizebox{\linewidth}{!}{

        \begin{tabular}{cccccccc} 
            \toprule
             Setting&Portion (clean/poison)   & BRISQUE & SDS & $IMS_{\text{CLIP}}$ & $IMS_{\text{VGG}}$ & CLIP-IQA & CLIP-IQA-C  \\ 
            \midrule
            \multirow{5}{*}{
                 \begin{tabular}[c]{@{}c@{}}Stand.\\Training\end{tabular}
            }
            &Clean                    &    14.610 $\pm$ 4.560           & 0.958 $\pm$ 0.060          & 0.781 $\pm$ 0.072          & 0.314 $\pm$ 0.427           & 0.818 $\pm$ 0.045          & 0.397 $\pm$ 0.113            \\
            &Mostly Clean(3/1)        & 14.700 $\pm$ 10.405 & 0.928 $\pm$ 0.047 & 0.785 $\pm$ 0.042 & 0.460 $\pm$ 0.325 & 0.799 $\pm$ 0.109 & 0.410 $\pm$ 0.203\\

            &Half-and-half (2/2)      &  16.123 $\pm$ 9.162 & 0.897 $\pm$ 0.103 & 0.785 $\pm$ 0.029 & 0.362 $\pm$ 0.315 & 0.733 $\pm$ 0.093 & 0.267 $\pm$ 0.191\\

           &Mostly Poison(1/3) &   17.801 $\pm$ 5.931 & 0.794 $\pm$ 0.121 & 0.761 $\pm$ 0.063 & 0.225 $\pm$ 0.468 & 0.670 $\pm$ 0.061 & 0.113 $\pm$ 0.110\\

            &Fully poisoned (4/0)  & 19.868 $\pm$ 2.051 & 0.068 $\pm$ 0.116 & 0.581 $\pm$ 0.044 & -0.299 $\pm$ 0.640 & 0.360 $\pm$ 0.085 & -0.520 $\pm$ 0.119\\
\midrule 

\multirow{5}{*}{
    \begin{tabular}[c]{@{}c@{}}Trans.\\Training\end{tabular}
}
&Clean                    &    19.063 $\pm$ 4.070          & 0.934 $\pm$ 0.092          & 0.756 $\pm$ 0.104          & 0.299 $\pm$ 0.357           & 0.750 $\pm$ 0.083          & 0.286 $\pm$ 0.042            \\

&Mostly Clean(3/1)        &  
24.385 $\pm$ 9.997 & 0.911 $\pm$ 0.077 & 0.794 $\pm$ 0.044 & 0.474 $\pm$ 0.216 & 0.763 $\pm$ 0.068 & 0.346 $\pm$ 0.129\\

&Half-and-half (2/2)      & 
25.809 $\pm$ 0.996 & 0.840 $\pm$ 0.062 & 0.769 $\pm$ 0.049 & 0.424 $\pm$ 0.194 & 0.715 $\pm$ 0.122 & 0.305 $\pm$ 0.247\\

&Mostly Poison(1/3) &   
23.588 $\pm$ 5.067 & 0.655 $\pm$ 0.299 & 0.728 $\pm$ 0.070 & 0.197 $\pm$ 0.499 & 0.592 $\pm$ 0.146 & 0.066 $\pm$ 0.312\\

&Fully poisoned (4/0)  & 12.982 $\pm$ 0.935 & 0.486 $\pm$ 0.156 & 0.668 $\pm$ 0.079 & -0.277 $\pm$ 0.636 & 0.534 $\pm$ 0.058 & -0.252 $\pm$ 0.030\\

            \bottomrule
            \end{tabular}
    }
    \caption{
        Performance of MetaCloak under low poisoning rate. The number in the portion column denotes the portion of clean images.
    }
    \label{tab: low-poisoning-rate.main.table}
\end{table*}

\begin{table*}[thbp]
    \centering
    \resizebox{\linewidth}{!}{
        \begin{tabular}{ccccccc} 
            \toprule
            Radius $r$ & BRISQUE  &SDS                       & $IMS_{\text{CLIP}}$         & $IMS_{\text{VGG}}$      & CLIP-IQAC                   & LIQE       \\
            \midrule
            Clean    & 21.783 {\scriptsize{$\pm$ 12.540}}  & 0.903 {\scriptsize{$\pm$ 0.291}}  & 0.790 {\scriptsize{$\pm$ 0.076}}  & 0.435 {\scriptsize{$\pm$ 0.657}}  & 0.329 {\scriptsize{$\pm$ 0.354}}  & 0.984 {\scriptsize{$\pm$ 0.124}} \\

    4/255  & 19.810 {\scriptsize{$\pm$ 8.828}}  & 0.832 {\scriptsize{$\pm$ 0.369}}  & 0.760 {\scriptsize{$\pm$ 0.094}}  & 0.207 {\scriptsize{$\pm$ 0.815}}  & -0.095 {\scriptsize{$\pm$ 0.318}}  & 0.883 {\scriptsize{$\pm$ 0.322}} \\

            8/255 & 17.271 {\scriptsize{$\pm$ 8.658}}  & 0.619 {\scriptsize{$\pm$ 0.480}}  & 0.705 {\scriptsize{$\pm$ 0.118}}  & 0.016 {\scriptsize{$\pm$ 0.856}}  & -0.278 {\scriptsize{$\pm$ 0.311}}  & 0.625 {\scriptsize{$\pm$ 0.484}} \\


            16/255 & 17.794 {\scriptsize{$\pm$ 6.707}}  & 0.393 {\scriptsize{$\pm$ 0.476}}  & 0.668 {\scriptsize{$\pm$ 0.077}}  & 0.067 {\scriptsize{$\pm$ 0.842}}  & -0.448 {\scriptsize{$\pm$ 0.203}}  & 0.359 {\scriptsize{$\pm$ 0.480}} \\

 32/255 & 16.962 {\scriptsize{$\pm$ 5.707}}  & 0.163 {\scriptsize{$\pm$ 0.367}}  & 0.628 {\scriptsize{$\pm$ 0.082}}  & -0.081 {\scriptsize{$\pm$ 0.851}}  & -0.532 {\scriptsize{$\pm$ 0.178}}  & 0.266 {\scriptsize{$\pm$ 0.442}} \\

            \bottomrule
            \end{tabular}
    }
    \caption{
        Performance of MetaCloak under the Trans. Training setting with different perturbation radii. 
    }
    \label{tab:radius.main.table}
\end{table*}

\begin{figure*}
   \centering
    \includegraphics[width=\linewidth]{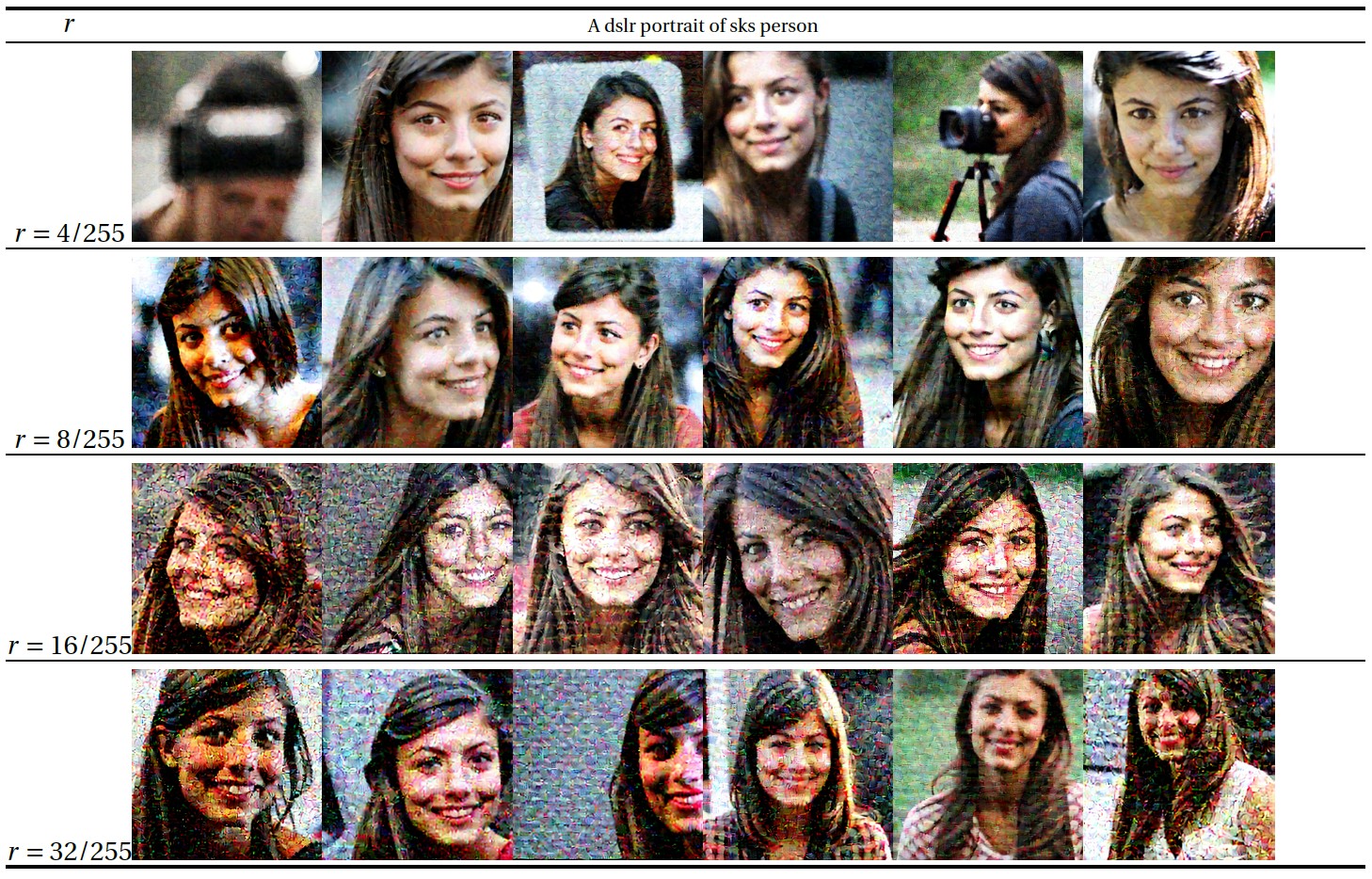}
  \caption{
      Performance of MetaCloak under different perturbation radii under Trans. Training setting. 
  }
  \label{fig: vis-diff-radius-gen}
\end{figure*}

\begin{figure*}[htbp]

    \centering
    \includegraphics[width=\linewidth]{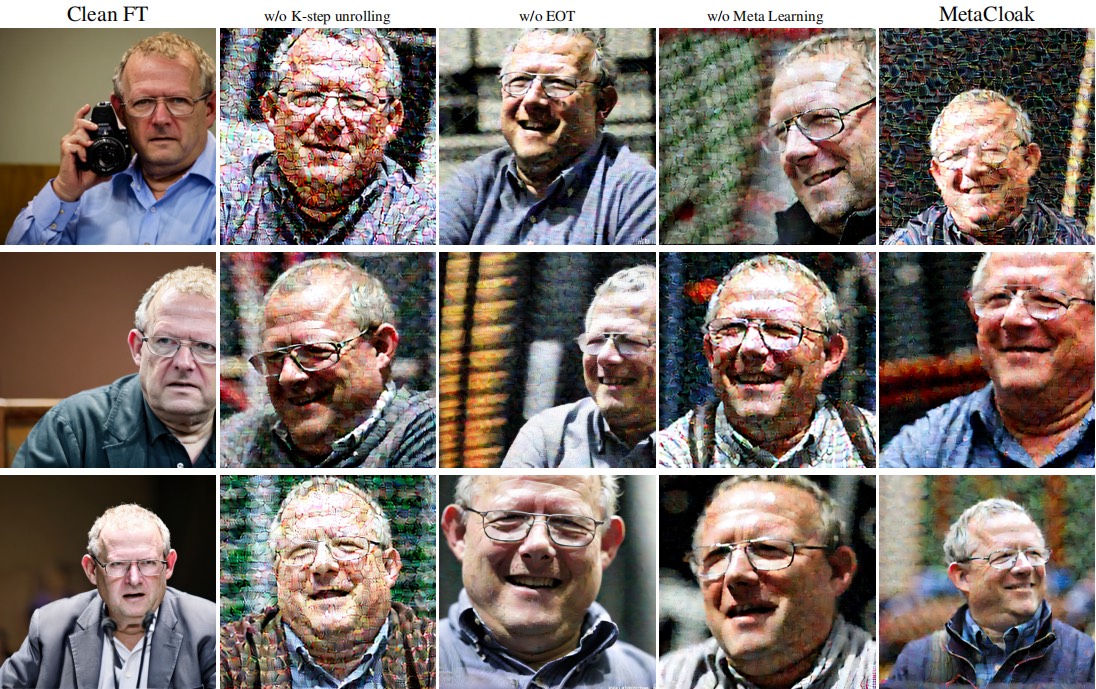}
    \caption{
    Effect of each component in MetaCloak on the quality of the generated images. The inferring prompt is ``a dslr photo of a person''. 
     }
    \label{fig: abl-vis-2}
\end{figure*}

%% file: tab/main-std-train.tex
\begin{table*}[htbp]
    \centering
    \resizebox{\linewidth}{!}{
        \begin{tabular}{ccccccc} 
            \toprule
              Dataset            & Method    &SDS $\downarrow$                       & $IMS_{\text{CLIP}}$    $\downarrow$      & $IMS_{\text{VGGNet}}$  $\downarrow$     & CLIP-IQAC $\downarrow$                   & LIQE $\downarrow$       \\
            \midrule
           \multirow{7}{*}{VGGFace2} & Clean       
           & 0.897 {\scriptsize{$\pm$ 0.302}}  & 0.814 {\scriptsize{$\pm$ 0.075}}  & 0.438 {\scriptsize{$\pm$ 0.658}}  & 0.456 {\scriptsize{$\pm$ 0.348}}  & 0.992 {\scriptsize{$\pm$ 0.088}} \\

                                          & ASPL      & 0.341 {\scriptsize{$\pm$ 0.471}}  & 0.607 {\scriptsize{$\pm$ 0.083}}  & -0.342 {\scriptsize{$\pm$ 0.822}}  & -0.457 {\scriptsize{$\pm$ 0.208}}  & 0.352 {\scriptsize{$\pm$ 0.477}} \\

                                                       & EASPL     & 0.356 {\scriptsize{$\pm$ 0.475}}  & \textbf{0.586} {\scriptsize{$\pm$ 0.098}}  & \textbf{-0.455} {\scriptsize{$\pm$ 0.780}}  & \textbf{-0.489} {\scriptsize{$\pm$ 0.223}}  & 0.219 {\scriptsize{$\pm$ 0.413}} \\

                                                          & FSMG       & 0.423 {\scriptsize{$\pm$ 0.488}}  & 0.611 {\scriptsize{$\pm$ 0.077}}  & -0.234 {\scriptsize{$\pm$ 0.843}}  & -0.402 {\scriptsize{$\pm$ 0.213}}  & 0.312 {\scriptsize{$\pm$ 0.464}} \\

                                                           & AdvDM      & 0.933 {\scriptsize{$\pm$ 0.241}}  & 0.674 {\scriptsize{$\pm$ 0.081}}  & 0.111 {\scriptsize{$\pm$ 0.821}}  & -0.177 {\scriptsize{$\pm$ 0.253}}  & 0.898 {\scriptsize{$\pm$ 0.302}} \\
 &Glaze  & 0.966 {\scriptsize{$\pm$ 0.174}}  & 0.762 {\scriptsize{$\pm$ 0.057}}  & 0.541 {\scriptsize{$\pm$ 0.544}}  & 0.012 {\scriptsize{$\pm$ 0.262}}  & 0.992 {\scriptsize{$\pm$ 0.088}} \\
 &PhotoGuard   & 0.967 {\scriptsize{$\pm$ 0.174}}  & 0.791 {\scriptsize{$\pm$ 0.064}}  & 0.548 {\scriptsize{$\pm$ 0.524}}  & 0.243 {\scriptsize{$\pm$ 0.284}}  & 1.000 {\scriptsize{$\pm$ 0.000}} \\

                                                       & MetaCloak   & \textbf{0.296} {\scriptsize{$\pm$ 0.448}}  & 0.662 {\scriptsize{$\pm$ 0.073}}  & -0.051 {\scriptsize{$\pm$ 0.838}}  & -0.380 {\scriptsize{$\pm$ 0.256}}  & \textbf{0.180} {\scriptsize{$\pm$ 0.384}} \\

            \midrule
              \multirow{7}{*}{CelebA-HQ}  & Clean      & 0.810 {\scriptsize{$\pm$ 0.389}}  & 0.763 {\scriptsize{$\pm$ 0.119}}  & 0.181 {\scriptsize{$\pm$ 0.784}}  & 0.470 {\scriptsize{$\pm$ 0.264}}  & 0.984 {\scriptsize{$\pm$ 0.124}} \\

                                                       & ASPL       & 0.809 {\scriptsize{$\pm$ 0.389}}  & 0.669 {\scriptsize{$\pm$ 0.079}}  & -0.238 {\scriptsize{$\pm$ 0.825}}  & -0.195 {\scriptsize{$\pm$ 0.275}}  & 0.883 {\scriptsize{$\pm$ 0.322}} \\

                                                              & EASPL     & 0.761 {\scriptsize{$\pm$ 0.421}}  & 0.656 {\scriptsize{$\pm$ 0.066}}  & -0.346 {\scriptsize{$\pm$ 0.817}}  & -0.226 {\scriptsize{$\pm$ 0.268}}  & 0.867 {\scriptsize{$\pm$ 0.339}} \\

                                                        & FSMG       & 0.684 {\scriptsize{$\pm$ 0.462}}  & \textbf{0.654} {\scriptsize{$\pm$ 0.084}}  & -0.445 {\scriptsize{$\pm$ 0.781}}  & -0.169 {\scriptsize{$\pm$ 0.255}}  & 0.773 {\scriptsize{$\pm$ 0.419}} \\

                                                            & AdvDM      & 0.849 {\scriptsize{$\pm$ 0.354}}  & 0.760 {\scriptsize{$\pm$ 0.062}}  & 0.331 {\scriptsize{$\pm$ 0.673}}  & 0.323 {\scriptsize{$\pm$ 0.287}}  & 1.000 {\scriptsize{$\pm$ 0.000}} \\
&Glaze  & 0.864 {\scriptsize{$\pm$ 0.338}}  & 0.784 {\scriptsize{$\pm$ 0.097}}  & 0.230 {\scriptsize{$\pm$ 0.756}}  & 0.486 {\scriptsize{$\pm$ 0.214}}  & 0.992 {\scriptsize{$\pm$ 0.088}} \\

&PhotoGuard  & 0.967 {\scriptsize{$\pm$ 0.174}}  & 0.805 {\scriptsize{$\pm$ 0.072}}  & 0.471 {\scriptsize{$\pm$ 0.557}}  & 0.424 {\scriptsize{$\pm$ 0.265}}  & 0.992 {\scriptsize{$\pm$ 0.088}} \\

                                                          & MetaCloak   & \textbf{0.407 }{\scriptsize{$\pm$ 0.485}}  & 0.666 {\scriptsize{$\pm$ 0.071}}  & \textbf{-0.654} {\scriptsize{$\pm$ 0.670}}  & \textbf{-0.354} {\scriptsize{$\pm$ 0.191}}  & \textbf{0.406} {\scriptsize{$\pm$ 0.491}} \\

            \bottomrule
        \end{tabular}
    }
    \caption{
        Results of different methods under the \textit{Stand. Training} setting with the corresponding std (±) on VGGFace2 and CelebA-HQ. 
    }
    \label{tab:main-standard-train}
\end{table*}

%% file: sections/appendix/recolor.tex
\subsection{Improving  Stealthiness with ReColorAdv}
\label{app. recolor}
To further improve the stealthiness of our method, we explore replacing the $\ell_\infty$-norm constrained attack with ReColorAdv \cite{laidlaw2019functional}, which generates adversarial images by applying a single pixel-wise function $f_g$ with parameters $g$ to every color value on the LUV color space. To make the perturbation imperceptible, $f_g$ is bounded such that for each pixel $\mathbf{c}_\mathbf{i}$, $\Vert f_g(\mathbf{c}_\mathbf{i}) - \mathbf{c}_\mathbf{i}\Vert_\infty \leq \epsilon$, 
and $\epsilon$ is set as 0.06, 0.08 and 0.10 respectively to observe the method's performance under different restrictions. To enforce the bound, we optimize the parameters $g$ with PGD, projecting $g$ back to the $\epsilon$ ball after every gradient step.

\begin{figure*}
    \centering
    \includegraphics[width=\linewidth]{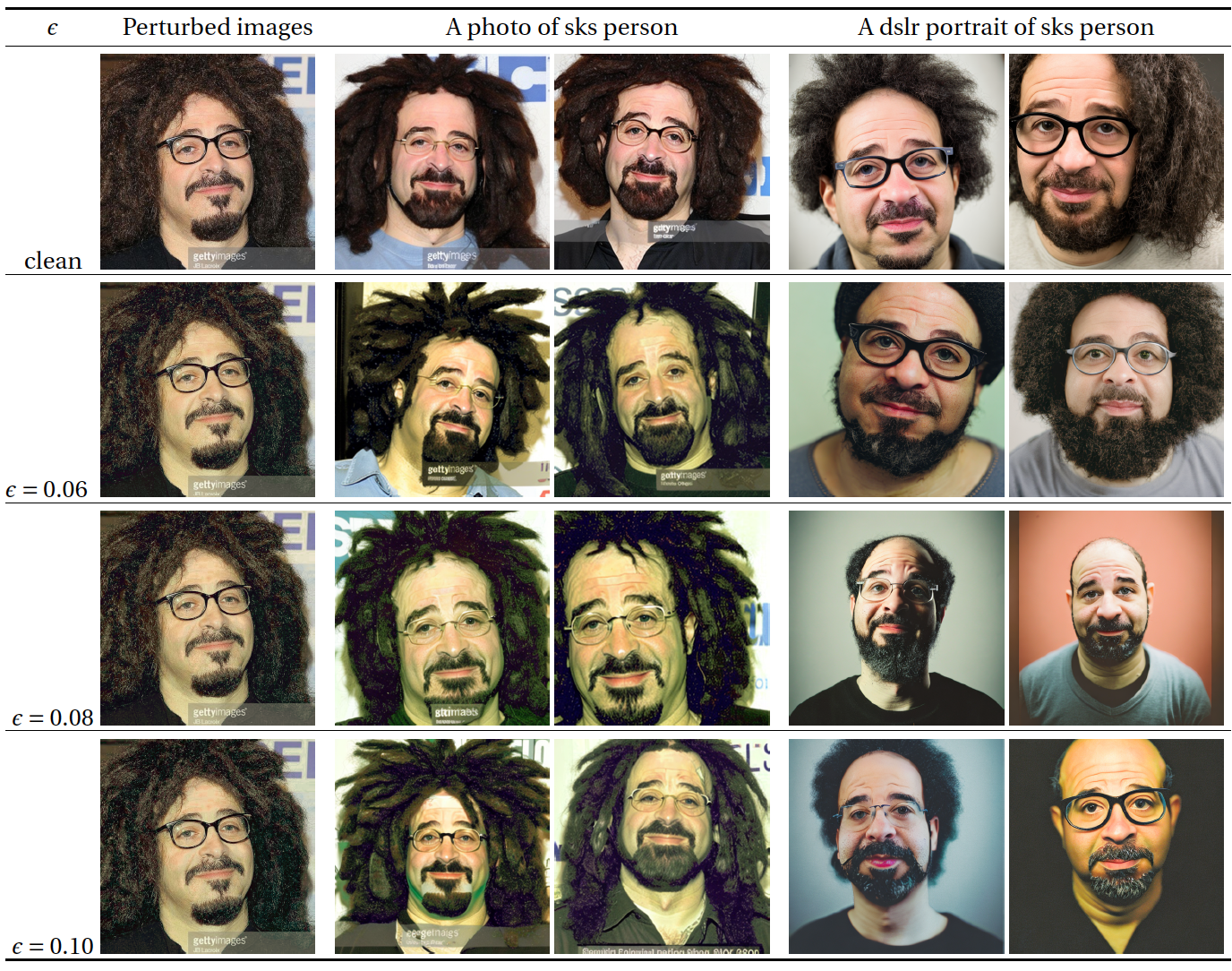}
    \caption{
        The Perturbed images by ReColorAdv and the generated images of DreamBooth trained on perturbed data. 
    }
    \label{fig:recolor}
\end{figure*}

The results are shown in Fig. \ref{fig:recolor}. As we can see from the figure, ReColorAdv crafts more uniform and global perturbations, which preserves dependencies between features such as the relationship between light and shadow and the shape boundaries. Unlike the $\ell_\infty$-norm attack, the defense effect of ReColorAdv is reflected in the odd color of the generated images instead of collapsed patterns and strips. However, ReColorAdv becomes less effective when the inferring prompt is different from the prompt used in the perturbation crafting process, which is "A photo of sks person" in our case.
In conclusion, we show that the ReColorAdv is promising to improve the stealthiness of our method further.

%% file: sections/appendix/understanding.tex
\subsection{Understanding Why MetaCloak works}
\label{app. understand}
\begin{figure*}
    \centering
    \includegraphics[width=0.6\linewidth]{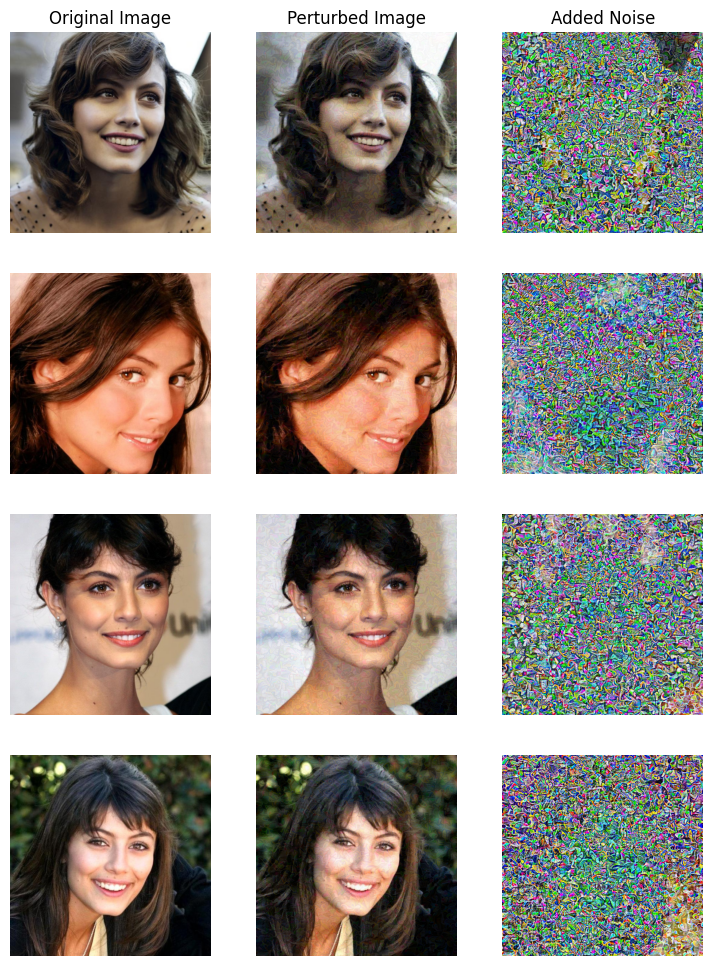}
    \caption{The original images, perturbed images, and the corresponding noise generated by MetaCloak on a sampled instance on the VGGFace2 dataset. }
    \label{fig:vis-metaclok}
\end{figure*}

\begin{figure*}[thbp]
    \centering
    \begin{subfigure}[b]{\textwidth}
        \centering
        \includegraphics[width=\textwidth]{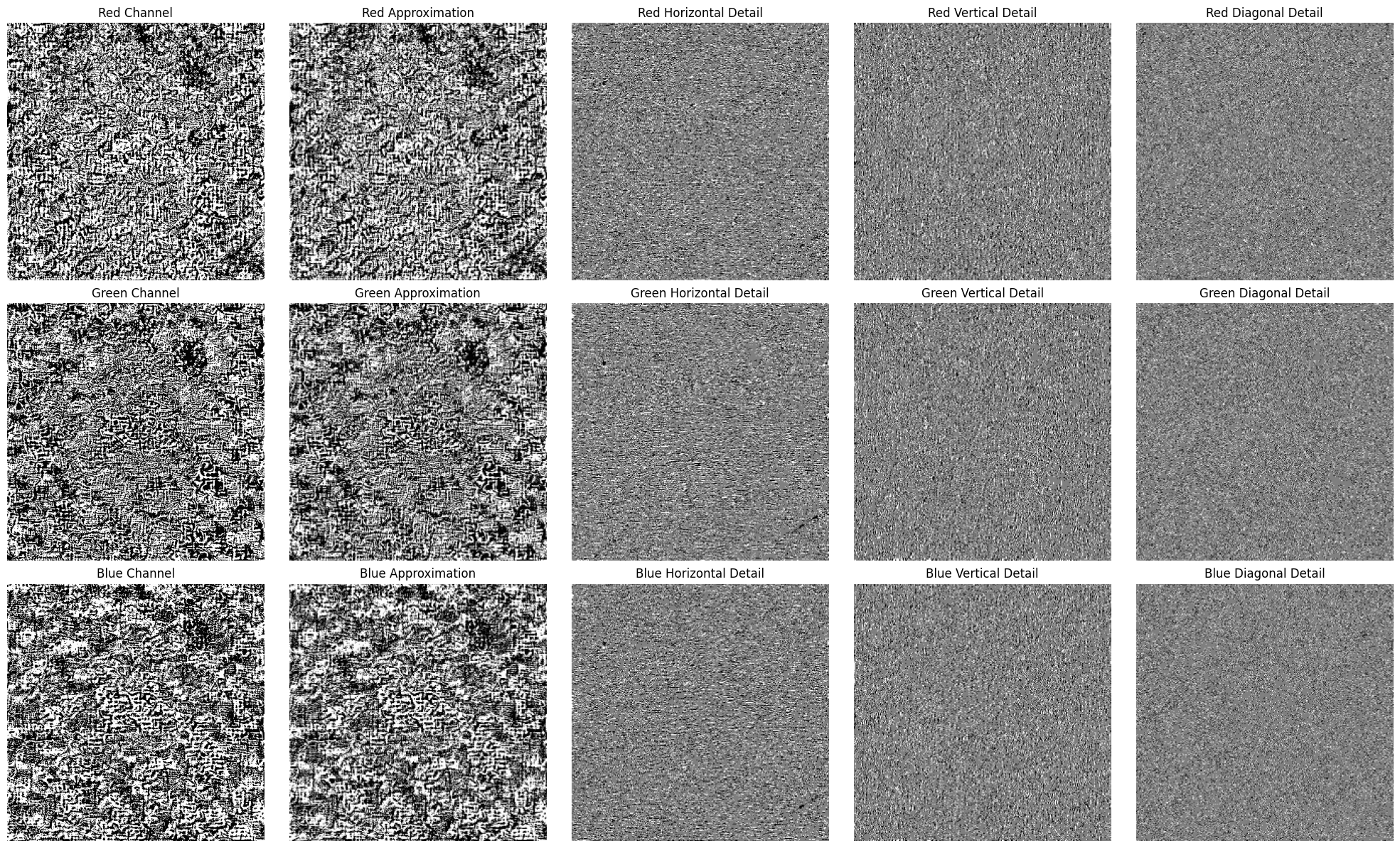}
        \caption{ASPL}
        \label{fig:under-aspl}
    \end{subfigure}
    \\ 
    \begin{subfigure}[b]{\textwidth}
        \centering
        \includegraphics[width=\textwidth]{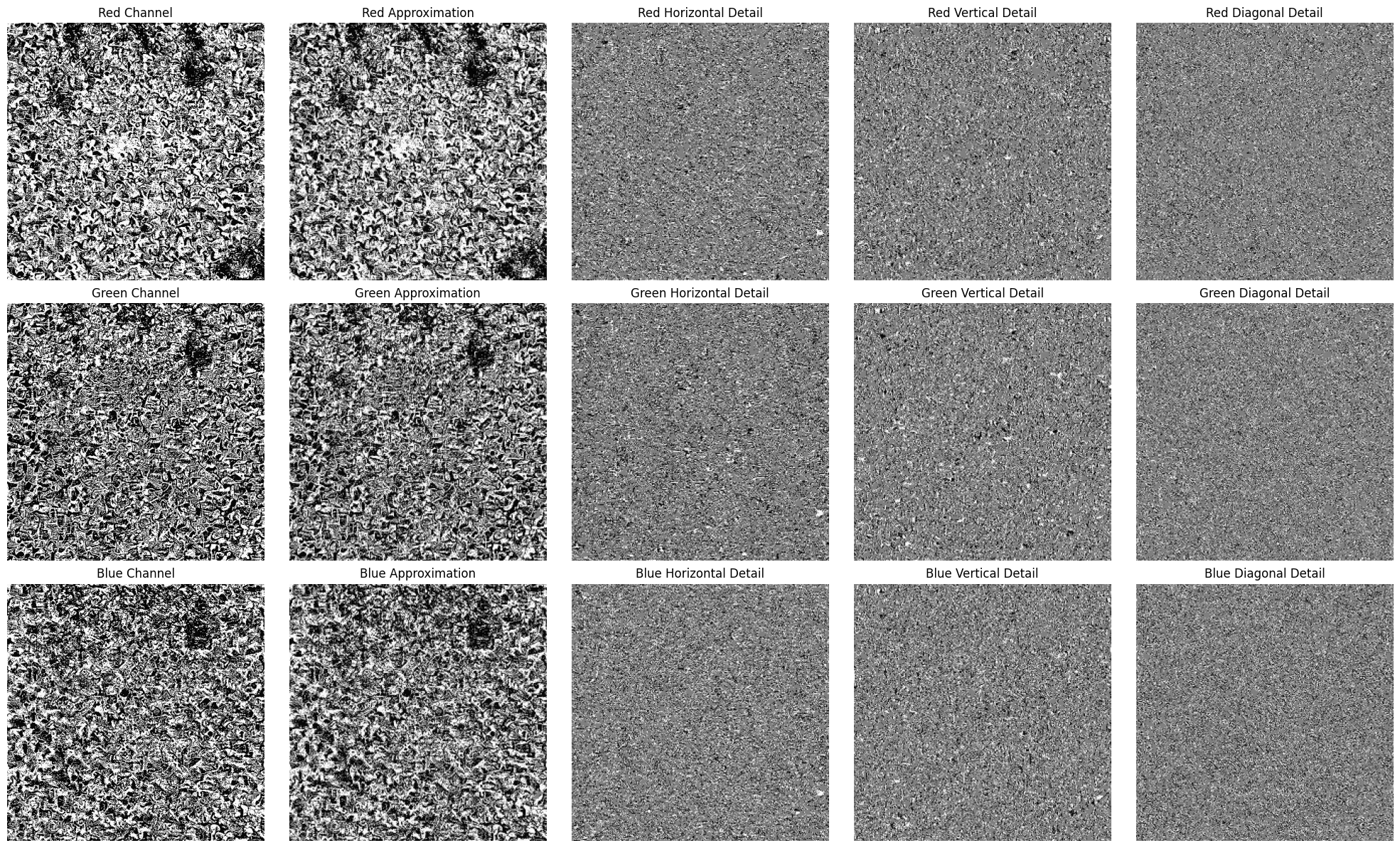}
        \caption{MetaCloak}
        \label{fig:under-metacloak}
    \end{subfigure}
    \caption{Comparison of wavelet decomposition of noise generated by ASPL and MetaCloak.}
    \label{fig:understanding-comparison}
\end{figure*}

\begin{figure*}
    \centering
    \includegraphics[width=\linewidth]{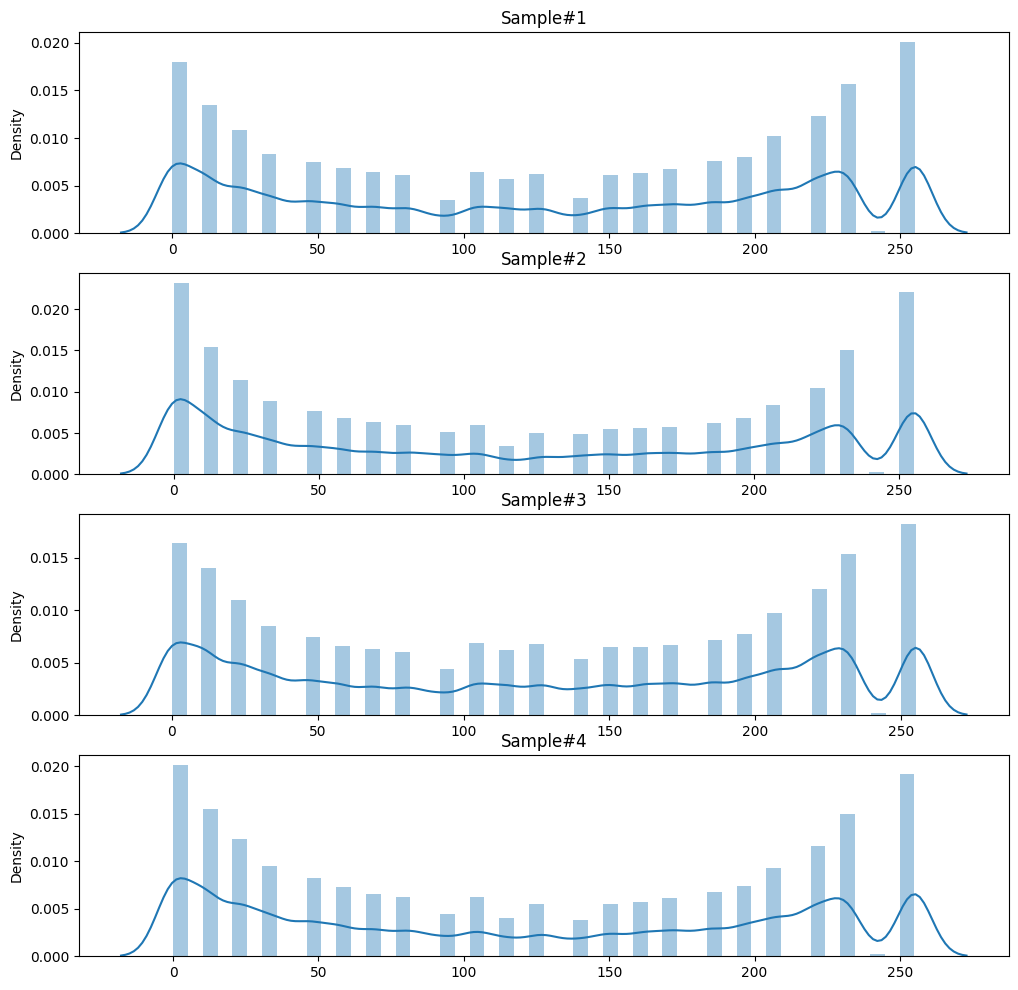}
    \caption{The density of noise scale generated by MetaCloak. We re-scale the pixel value of noise to the range $[0, 255]$ for visualization. 
    }
    \label{fig:understanding-density}
\end{figure*}

To understand why MetaCloak works, we visualize the learned noise and conduct wavelet analysis on the noise. We first visualize the original images, perturbed images, and the corresponding noise in Fig.~\ref{fig:vis-metaclok}. As can be seen from Fig.~\ref{fig:understanding-comparison}, the noise learned by MetaCloak is significantly sharper than the noise generated by the previous method. Thus, the pattern of our noise is less likely to be obscured and turn blurry when encountering defensive measures such as Gaussian transformation and is more likely to fool subject-driven generative models such as Dreambooth.
We further plot the distribution of the scales of the learned noise in Fig.~\ref{fig:understanding-density}. As can be observed, the scales of the noise are polarized, clustering at both extreme values, and thus more resistant to perturbations.

